\documentclass[10pt,letterpaper]{article}
\usepackage[top=0.85in,left=2.75in,footskip=0.75in]{geometry}

\usepackage{amsmath,amssymb}

\usepackage{changepage}

\usepackage[utf8x]{inputenc}

\usepackage{textcomp,marvosym}

\usepackage{cite}

\usepackage{nameref,hyperref}

\usepackage[right]{lineno}

\usepackage{microtype}
\DisableLigatures[f]{encoding = *, family = * }

\usepackage[table]{xcolor}

\usepackage{array}

\newcolumntype{+}{!{\vrule width 2pt}}

\newlength\savedwidth

\raggedright
\usepackage[aboveskip=1pt,labelfont=bf,labelsep=period,justification=raggedright,singlelinecheck=off]{caption}

\makeatletter
\renewcommand{\@biblabel}[1]{\quad#1.}
\makeatother

\DeclareCaptionLabelFormat{plain}{#2} 

\usepackage{lastpage,fancyhdr,graphicx}
\usepackage{epstopdf}
\fancyheadoffset[L]{2.25in}
\fancyfootoffset[L]{2.25in}
\usepackage{xcolor}
\usepackage[ruled,vlined]{algorithm2e}

\SetCommentSty{mycommfont}

\SetKwInput{KwInput}{Input}                
\SetKwInput{KwOutput}{Output}              

\begin{document}
\vspace*{0.2in}

\begin{flushleft}
{\Large
\textbf\newline{Towards minimizing efforts for Morphing Attacks -- Deep embeddings for morphing pair selection and improved Morphing Attack Detection} 
}
\newline
\\
Roman Kessler\textsuperscript{1,2,*},
Kiran Raja\textsuperscript{1},
Juan Tapia\textsuperscript{2},
Christoph Busch\textsuperscript{1,2}
\\
\bigskip
\textbf{1} Norwegian University of Science and Technology (NTNU), Gjøvik, Norway
\\
\textbf{2} Hochschule Darmstadt, Darmstadt, Germany
\\

\bigskip

%
%





* Corresponding author\\
E-mail: rkesslerx@gmail.com (RK)

\end{flushleft}
\section*{Abstract}

Face Morphing Attacks pose a threat to the security of identity documents, especially with respect to a subsequent access control process, because they allow both involved individuals to use the same document. Several algorithms are currently being developed to detect Morphing Attacks, often requiring large data sets of morphed face images for training. 

In the present study, face embeddings are used for two different purposes: first, to pre-select images for the subsequent large-scale generation of Morphing Attacks, and second, to detect potential Morphing Attacks. Previous studies have demonstrated the power of embeddings in both use cases. However, we aim to build on these studies by adding the more powerful MagFace model to both use cases, and by performing comprehensive analyses of the role of embeddings in pre-selection and attack detection in terms of the vulnerability of face recognition systems and attack detection algorithms. In particular, we use recent developments to assess the attack potential, but also investigate the influence of morphing algorithms.

For the first objective, an algorithm is developed that pairs individuals based on the similarity of their face embeddings. Different state-of-the-art face recognition systems are used to extract embeddings in order to pre-select the face images and different morphing algorithms are used to fuse the face images. The attack potential of the differently generated morphed face images will be quantified to compare the usability of the embeddings for automatically generating a large number of successful Morphing Attacks.
For the second objective, we compare the performance of the embeddings of two state-of-the-art face recognition systems with respect to their ability to detect morphed face images. 

Our results demonstrate that ArcFace and MagFace provide valuable face embeddings for image pre-selection. Various open-source and commercial-off-the-shelf face recognition systems are vulnerable to the generated Morphing Attacks, and their vulnerability increases when image pre-selection is based on embeddings compared to random pairing. In particular, landmark-based closed-source morphing algorithms generate attacks that pose a high risk to any tested face recognition system. Remarkably, more accurate face recognition systems show a higher vulnerability to Morphing Attacks. Among the systems tested, commercial-off-the-shelf systems were the most vulnerable to Morphing Attacks.
In addition, MagFace embeddings stand out as a robust alternative for detecting morphed face images compared to the previously used ArcFace embeddings.

The results endorse the benefits of face embeddings for more effective image pre-selection for face morphing and for more accurate detection of morphed face images, as demonstrated by extensive analysis of various designed attacks. The MagFace model is a powerful alternative to the often-used ArcFace model in detecting attacks and can increase performance depending on the use case.
It also highlights the usability of embeddings to generate large-scale morphed face databases for various purposes, such as training Morphing Attack Detection algorithms as a countermeasure against attacks.


\section*{Introduction}
\label{sec:introduction}

Automated face recognition plays an integral role in access control, criminal investigation, and surveillance settings \cite{kortli_face_2020}. In particular, for automated border control, the observation and analysis of facial characteristics is becoming increasingly important for identity verification \cite{sanchez_del_rio_automated_2016, noauthor_regulation_2017}. For example, to assist immigration officers at borders or airports, automated Facial Recognition Systems (FRS) can increase traveler throughput and reduce costs.

In a typical identity verification process, a biometric reference image, i.e., a passport photograph of a subject is compared to one or multiple probe images, i.e., trusted live photographs captured at the border. A similarity score is then calculated between the reference and probe images, and the subject is allowed to cross the border if the similarity score exceeds a predetermined threshold. 

\begin{figure}[!h]
 \centering
 \includegraphics[width=1.\linewidth]{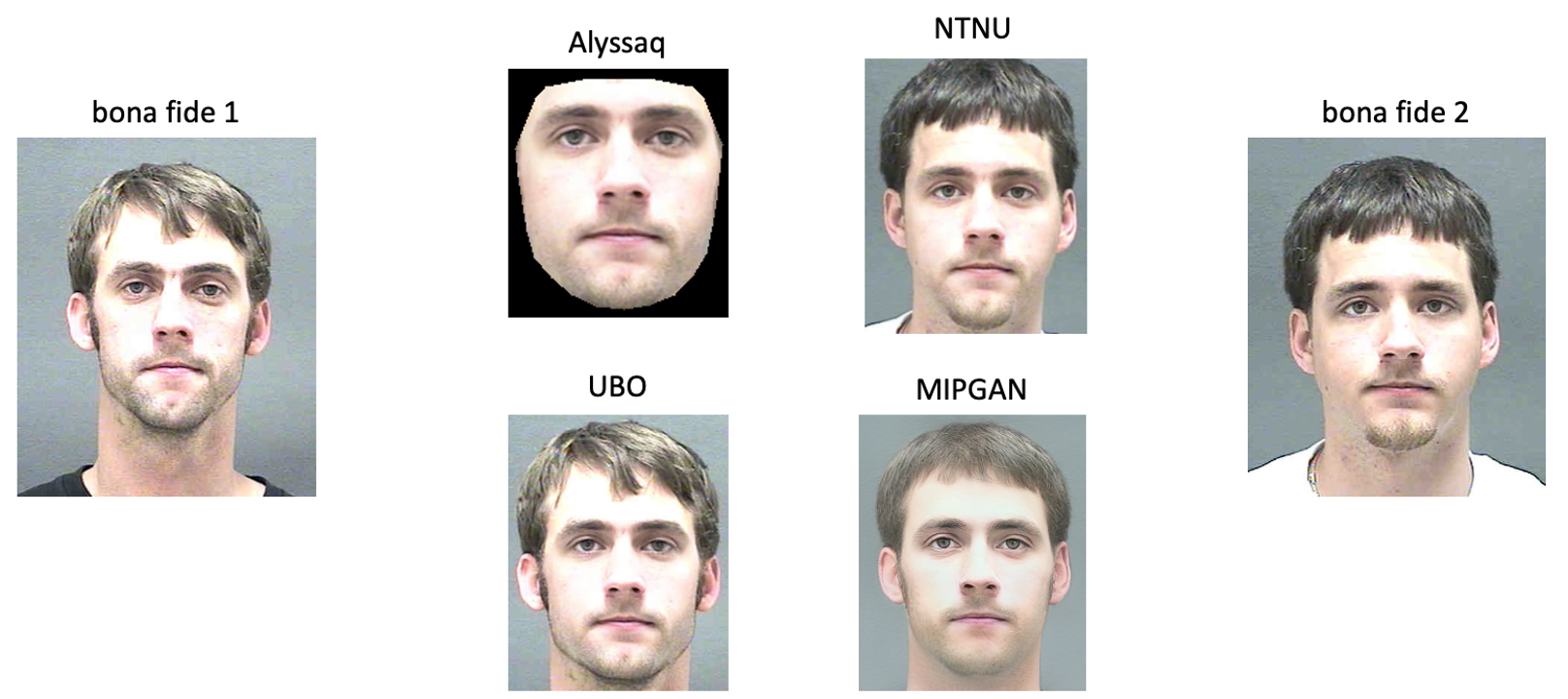}
 \caption{{\bf Illustration of morphed face images created using different morphing approaches.}
  The images on the left and on the right represent the corresponding two bona fide face images.\\ \emph{Face images are republished from \cite{uncw_morph_nodate} under a CC BY license, with permission from Prof. Karl Ricanek Jr, University of North Carolina at Wilmington, original copyright 2006.}}
 \label{fig:examplemorphs}
\end{figure}

The operation of such an automated FRS must be secure and robust. However, so-called Morphing Attacks can compromise the security of FRSs \cite{scherhag_face_2016, Ferrara2016}. In a Morphing Attack, an attacker combines the face images of two or more subjects in order to form a morphed face image (see e.g., Fig~\ref{fig:examplemorphs}). This morphed face image is presented as a (manipulated) reference to the FRS as it is stored and read from the passport on request. Since the calculated similarity score between the morphed face reference image and one or more bona fide probe images should be high enough to exceed some predetermined decision threshold $\tau$ of the FRS, each attacker's identity is falsely verified. As a result, two or more individuals may use the same passport to cross the border, and the unique link between a passport and an individual is broken.

Real-world Morphing Attack cases have already been reported (e.g., \cite{torkar_morphing_2021}). High-ranking governmental bodies, such as EU DG HOME and the Ministries of Interior of the G7 states have now formed an action, in order to address the Morphing Attack Detection topic.

In the recent years, Morphing Attack Detection (MAD) algorithms have been proposed to fend off such attacks \cite{venkatesh_face_2021}. Several MAD algorithms are based on machine learning and therefore require a large amount of data for training (e.g., \cite{ferrara_face_2021, tapia2021}). However, generating such a large data source with high quality morphs is often challenged by the need for manual post-processing to reduce image artifacts \cite{scherhag_face_2016, scherhag_biometric_2017}. It is therefore important to develop criteria that allow an informed but automated selection of two (or more) individuals suitable for producing a high quality morph image \cite{roettcher_finding_2020} without relying heavily on manual intervention. These criteria can then be used to find a large number of possible pairs of suitable source images from which morphs can be automatically generated, and a database of morphed images can be created for future research on MAD. 

Previous research has shown that an adequate pre-selection of possible morph pairs can reduce two things: (i) the choice of the applied morphing algorithm is less relevant \cite{roettcher_finding_2020}, and (ii) the amount of artifacts produced by an automated morphing algorithm is reduced, making an FRS more vulnerable to the Morphing Attack \cite{roettcher_finding_2020}. A large database of morphed images not only allows for better training and testing of MAD algorithms. It also allows for statistical analysis of the performance of FRSs, and may ultimately lead to a better understanding of the image properties that predict the success of a Morphing Attack. We claim that our analysis will contribute to the creation of large-scale training data sets to make MAD approaches more robust.

We further note that manual image pre-selection relies on some heuristic criteria as employed in previous works \cite{raghavendra_face_2017, raja_morphing_2020, scherhag_biometric_2017}. For instance, soft biometrics characteristics have been used, to morph only subjects of similar age, same gender, or same ethnicity \cite{raghavendra_face_2017, raja_morphing_2020, scherhag_biometric_2017}. In a complementary direction, other characteristics such as the shape of the hair, skin tone, differences in landmark position, and Euclidean distance between \emph{face embeddings} extracted from the OpenFace model \cite{amos_openface_nodate} have shown positive effects on the attack potential of a morph \cite{roettcher_finding_2020}.

Deep learning-based FRSs provide feature embeddings, which are low-dimensional representations of high-dimensional face images \cite{will_koehrsen_neural_2018}. In the context of face recognition, termed \emph{face embeddings}, feature embeddings are point representations in latent space learned during the training of a face recognition neural network \cite{shi_probabilistic_2019}. Computing a simple distance in latent space between two face embeddings, such as the cosine distance, can be effective in quantifying the similarity of two faces \cite{schroff_facenet_2015}. Motivated by the superior performance of models that use such embeddings for face recognition (e.g., \cite{deng_arcface_2019-1, meng_magface_2021, amos_openface_nodate}), we hypothesize that feature embeddings from deeply learnt face models can provide rich enough data to automate image pre-selection for morphing simply by analyzing the embeddings. 

We take advantage of the power of embeddings in determining similarity by presenting them as auxiliary data for image pre-selection in morphing. The general assumption in our work is that a small distance between the face embeddings of two subjects corresponds to a high similarity (structural and perceptual) of the facial features of the two subjects. Thus, selecting pairs of face images based on high similarity scores between them can help generate more realistic morphs compared to selecting two face images that do not look particularly similar. Automating the pair selection process (i.e., pre-selection) makes it tractable, reproducible, scalable, and less subjective than manual approaches.

An attacker could also use embeddings by comparing a number of candidate image embeddings to find a suitable morphing partner, for example in a database of possible accomplices. This could improve the success of an attack by allowing more quantifiable parameters to be used in deciding which morphing partner to choose, rather than just soft biometrics and subjective facial similarity. 

From a theoretical standpoint, it is obvious that pre-selection based on face similarity can increase the attack potential of resulting morphs \cite{gomez-barrero_is_2017, scherhag_biometric_2017}, and previous research has demonstrated the increased attack potential of pre-selected morphs \cite{roettcher_finding_2020} using OpenFace \cite{amos_openface_nodate} embeddings. However, a detailed analysis of morphing pairs pre-selected on embeddings of different state-of-the-art FRSs such as ArcFace and MagFace is still lacking. Insights on the suitability of these contemporary models for image pre-selection are however crucial to guide future attempts to create large-scale databases of morphed face images, which are especially important for the research context of MAD.

We evaluate the pre-selections by quantifying the attack potential of the created morphs on different FRSs. In addition to previously deployed metrics, we also use the recently introduced Morphing Attack Potential (MAP) \cite{ferraraMorphVulnerabilityRate2022, isoiec_international_2023} and a few of its derivatives. MAP compensates for some drawbacks of earlier metrics. E.g., the MMPMR \cite{scherhag_face_2016} tends to represent the upper bound of attack potential, since an attack is considered successful with only one (of several) bona fide images positively verified against the morphed reference. On the other hand, the FMMPMR  \cite{venkatesh_influence_2020} represents the lower limit of the attacker potential, since an attack here is considered a success only in case of exclusively positive verification of all bona fide face images. However, the bona fide face images in a real-world attack are often very similar among themselves, as they are all captured in short time intervals just before the verification process. In contrast, the face images used in the present study were taken with large time intervals between them, and are therefore considerably more heterogeneous, so FMMPMR is not a pertinent measure of attack potential. MAP, on the other hand, tests attack potential across several different FRSs and thus provides a more generalizable picture of the actual attack potential. In real-life scenarios, attackers often do not know which system is being used for verification. Therefore, different attacks are launched on each system. If the bona fide face images are as heterogeneous as in the present study, this measure is also characterized by greater robustness.

We further make use of the learnt embeddings by using a particular category of losses. The magnitude of these losses can measure the quality of the given face image. We use these losses for creating a robust MAD algorithm in a Differential-MAD (D-MAD) setting. Our D-MAD algorithm is further based on the idea that the magnitude of the feature embedding is highly correlating and monotonically increasing if the pair of images are to be chosen using FRSs with adaptively learnt intra-class and inter-class feature distributions \cite{meng_magface_2021}.

For this D-MAD algorithm, we deliberately build upon the concept of a previously published D-MAD algorithm, which used ArcFace embeddings \cite{scherhag_deep_2020}. However, we take advantage of the better face recognition performance of MagFace \cite{meng_magface_2021} and therefore train a D-MAD classifier on the differential face embeddings of this FRS, while building an identical D-MAD algorithm using ArcFace embeddings for comparison.

The present study makes the following contributions:

\begin{itemize}
    
    \item We first examine face embeddings produced by several well-known FRSs for automated image pre-selection to produce morphed face images. We demonstrate that a large data set of morphed face images can be easily constructed by analyzing the distance between the embeddings. We empirically validate the effectiveness of our developed selection criteria by systematically studying the susceptibility of deep learning-based FRSs and two commercial-off-the-shelf (COTS) FRSs on our generated database.
    
    \item We generalize our proposed pre-selection approach for morph generation across different morphing algorithms and validate it on several FRSs using multiple attack attempts. In particular, we compute the recently proposed Morphing Attack Potential (MAP) metric on the resulting data set of morphed images, which illustrates for a more generalizable and robust measure of attack potential. Our experiments show that pre-selection can produce better morphs and can compromise FRSs and MAD classifiers to a high degree, regardless of which particular embeddings were used in the pre-selection. We validate our pre-selection approach against a control data set consisting of randomly paired face images. However, the model used for preselection has a significant impact on the attack potential and efficiency to avoid detection by MAD classifiers.
     
    \item Furthermore, motivated by the limited performance of MAD algorithms in detecting Morphing Attacks generated by our pipeline, we present a newly designed MAD algorithm using MagFace over ArcFace differential embeddings for training, improving the detection capability. 
    
\end{itemize}

In the rest of the paper, we first present our proposed approach for morph pair pre-selection and provide details about the data sets and models used to provide embeddings, morph face images, and validate the resulting morphs. We then illustrate the results of the Morphing Attacks by showing how the attacks generated by our pipeline are able to fool FRSs. Finally, we construct and benchmark a new MAD algorithm on this data set and discuss our results.

\section*{Methods}

\subsection*{Proposed approach for morph pair pre-selection}
\label{methods:algo}

\begin{figure}[!h]
 \centering
 \includegraphics[width=1.\linewidth]{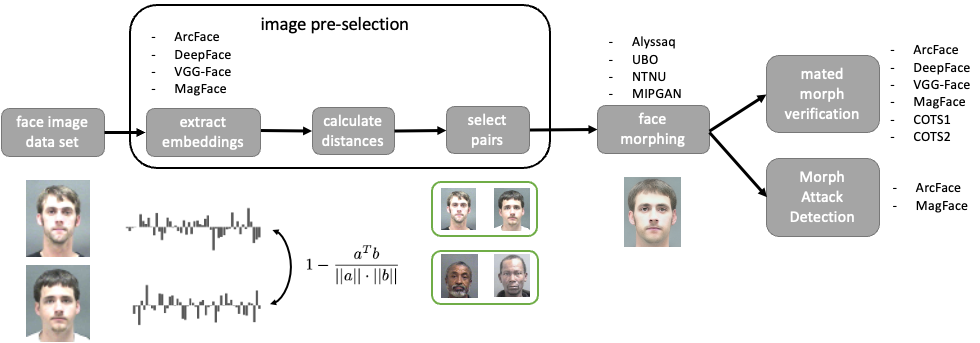} 
 \caption{{\bf General workflow of our proposed pipeline for image pre-selection.}
 Embeddings were extracted from one sample of each subject. Distances between embeddings were calculated. Faces were paired based on a low distance between embeddings. Pairs were then morphed, and morphed images were verified against bona fide probe images. Furthermore, Morphing Attack Detection has been conducted. The image pre-selection steps are further illustrated in Algorithm~\ref{algo1}. The processing steps were performed using different FRSs and different morphing algorithms.\\ \emph{Face images are republished from \cite{uncw_morph_nodate} under a CC BY license, with permission from Prof. Karl Ricanek Jr, University of North Carolina at Wilmington, original copyright 2006.}}
 \label{fig:track12}
\end{figure}

Our proposed approach consists of generic deep learning-based FRSs to extract embeddings followed by a similarity-based pair-selection module. The selected pairs were then provided to different face morphing algorithms. The generated data set is thereupon used to study vulnerability of FRSs and to develop a MAD algorithm. Fig~\ref{fig:track12} presents an illustration of our proposed approach for the convenience of the reader.

\subsubsection*{Embeddings from Face Recognition Systems} 
\label{sec:emb}

In our proposed architecture, different state-of-the-art implementations of FRSs were used to extract face embeddings for image pre-selection. Based on the results reported in recent work, we selected four different architectures to obtain the embeddings in our pre-selection pipeline. We chose ArcFace \cite{deng_arcface_2019}, VGG-Face \cite{parkhi_deep_2015}, DeepFace \cite{taigman_deepface_2014}, and MagFace \cite{meng_magface_2021}. For ArcFace, VGG-Face, and DeepFace (Facebook), Tensorflow implementations of the respective models were used, which were included in the software distribution of the LightFace repository \cite{serengil2020lightface}. For MagFace, the official repository was used \cite{meng_magface_2021}. Each of these FRSs provides an embedding vector representing a face image. The vector differs in length depending on which FRS was used, as shown in Table~\ref{table:numberemb}. For the sake of completeness of the experiments in this paper, we also used the same set of FRSs to verify the resulting morphed faces, in addition to two COTS FRSs.

\begin{table}[!t] \centering
\caption{{\bf The number of embeddings per FRS.}}
\label{table:numberemb}
\begin{tabular}{@{} c c @{}}     \hline
FRS & \# of embeddings \\  \hline
ArcFace & 512 \\ 
DeepFace & 4096 \\ 
VGG-Face & 2622 \\ 
MagFace & 512 \\
 \hline
\end{tabular}
\end{table}

\subsubsection*{Pre-selection algorithm} 

Our proposed pre-selection criterion is based on a measure of similarity of embeddings which typically consist of rich identity preserving information \cite{nguyen2010cosine}. Given two equally sized embedding vectors, we employed Cosine distance (Eq~\ref{eq:cosine}) to determine the similarity between the underlying faces. For a pair of embeddings, corresponding to two face images, the Cosine distance \cite{nguyen2010cosine} can be defined as:

        \begin{equation} \label{eq:cosine}
            d_{cos}(E_1,E_2) = 1 - \frac{E_1^T E_2}{||E_1||\cdot||E_2||}
        \end{equation}
 
with $E_1$ and $E_2$ the $D$-dimensional embedding vectors of the images.

For a given data subject and a chosen FRS, we computed the Cosine distances between the subject and the remaining $N$ subjects in the chosen database. The procedure was repeated $N$ times for each subject in the database. 
We further enforced a demographic consistency check. Specifically, we enforced that gender and ethnicity between individuals of a potential pair must correspond. Further, we allowed for a maximum age difference of $5$ years between individuals of a potential pair. The labels provided with the face image data set were used for these checks (see below).
Based on the computed similarity score matrix across all subjects, we retained the upper diagonal of the score matrix owing to the symmetric nature of the Cosine distance (i.e., $d_{cos}(E_1,E_2) = d_{cos}(E_2,E_1)$). The face images of unique subjects fulfilling the criteria were then chosen for morphing. The details of our pre-selection criteria are presented in Algorithm~\ref{algo1} for the sake of brevity. It should be mentioned that by this procedure each data subject was only used for the creation of a maximum of one morph pair. However, each data subject can be used in all runs of the algorithm, i.e., when the algorithm was applied to the embeddings of another FRS.

\begin{algorithm}[!ht]
\DontPrintSemicolon
  \KwInput{list of FRSs; face image data set}
  \KwOutput{lists of pairings ($1$ list for each FRS)}
  \KwData{face images ($1$ sample per data subject)}
   \tcc{loop over FRSs}
   \For{$f \in FRSs$}
   {
   	   \tcc{loop over different face images}
       \For{$i, I \in ind, images$}
           {
           		$I \longleftarrow preprocess(I)$ \tcp*{align \& crop}
           		$E_i \longleftarrow extract(I)$ \tcp*{extract embeddings}
           }
       
       \tcc{for all combinations of images}
       \For{$i, x \in index_1, E$}
           {
           \For{$j, y \in index_2, E$}
               {
               	\If{$i \geq j$}
                     {
                        $D_{i,j} \longleftarrow 1 - \frac{x^T y}{||x||\cdot||y||} $ \tcp*{calculate Cosine distance}
                    }
                    
                \Else 
                    {
                        $D_{i,j} \longleftarrow NaN$ \tcp*{use symmetry \& omit self-comparison}
                    }

               }
           }
       \tcc{pair the face images via embeddings}
        \While{$>1$ face image unpaired}
        {
          $i,j = argmin(D)$ \tcp*{find 2-D index of lowest value in      distance matrix}
          \If{$ | age_i - age_j | \leq 5 $ \& $ gender_i = gender_j $ \& $ ethnicity_i = ethnicity_j $}
            {
                $pairs(f).append(\{i,j\})$ \tcp*{create pair}
                $D(i,*) \longleftarrow NaN$ \;
                $D(*,j) \longleftarrow NaN$\tcp*{both subjects are excluded from further pairing}
            }
            \Else
            {   $D(i,j) \longleftarrow NaN$ \tcp*{only this pair is excluded from pairing, both subjects remain available for alternative pairings}
            }
        }
   }
\caption{Image pre-selection}
\label{algo1}
\end{algorithm}

Pair selection according to this algorithm was carried out using four different FRSs. The algorithm was reapplied to create a baseline comparison data set without taking into account the similarity between the embeddings. For the baseline data set, only the face images of subjects with matching demographics were randomly morphed.

\subsection*{Creation of the morphed face data set}  
\label{methods:rest}
  
We validate our proposed approach for image pre-selection using the academic version of the UNCW-MORPH face data set distributed by the face ageing group of the University of North Carolina Wilmington (UNCW) \cite{ricanek_morph_2006, uncw_morph_nodate}. UNCW-MORPH contains bona fide face images only, contrary to its name which is suggesting it contains morphed images. It comprises over $55,000$ face images of more than $13,000$ data subjects, captured between 2003 and 2007. The facial images were captured in frontal poses with largely neutral expressions, making the data set suitable for face morphing. The face images had resolutions between $200$ px $\times$ $240$ px and $400$ px $\times$ $480$ px, with each image labeled with exact age, gender, and ethnicity.

In order to employ the data for our experiments, we conducted a curation process with a number of pre-processing steps. First, all samples were checked for neutral facial expressions and any images not conforming to neutral expressions were eliminated. We specifically used an emotion detection model from the LightFace package \cite{serengil2020lightface} to verify neutral expression. All samples, for which \emph{neutral} was not the emotion with the highest probability, or samples, for which the emotion model failed, were discarded from morphing.

Since we needed multiple samples to study the susceptibility of FRSs to the resulting morphs, any subject with fewer than five samples was discarded from further analysis. For the remaining subjects, the first sample (in chronological order) was used for morphing, while the remaining samples were used for validation. Of the $55,134$ samples from $13,618$ subjects in the raw data set, $22,992$ samples from $3,337$ subjects remained in the data set.

\subsubsection*{Morphing algorithms}
To perform the morphing, we chose four different morphing algorithms. Three of them were landmark-based (Alyssaq morpher \cite{alyssa_quek_face_nodate}, NTNU morpher \cite{raghavendra_face_2017, raja_morphing_2020}, \& UBO morpher \cite{Ferrara2014,Ferrara2016,Ferrara2018,Ferrara2019}). In these landmark-based algorithms, morphing was based on averaging the landmark coordinates of the two morph candidate images. The $68$ face landmarks were extracted using the OpenCV dlib library \cite{kazemi_one_2014}, with an ensemble of regression trees used to estimate the coordinates. As a fourth morphing algorithm, a deep learning based algorithm -- Identity Prior driven Generative Adversarial Networks (MIPGAN) \cite{zhang_mipgan_2021, venkatesh_can_2020} -- were used. Unlike landmark morphed images, MIPGAN used the latent space of two samples to generate morphed images. A morphing factor (alpha) of $0.5$ was used for all morphing algorithms. No image pre-processing or post-processing was done other than the steps included in the respective morphing packages. However, rescaling and cropping steps were performed for the face recognition steps in the vulnerability analysis (see below). Fig~\ref{fig:examplemorphs} illustrates exemplary morphed images created by all four approaches.

\subsubsection*{Properties of the morphed images}
Morphing was then performed based on several criteria, such as the similarity of facial embedding vectors and demographic consistency. Since several different FRSs were used for the pre-selection algorithm, the pairings naturally differed between the approaches used. Furthermore, there were differences in the absolute number of pairs found. This is due to the fact that the demographic check eliminates many matched pairs if the demographic properties are too different. Thus, we obtained $452$ pairs when pre-selecting with ArcFace, $511$ for DeepFace, $632$ for VGG-Face, and $639$ for MagFace. When random pairing was performed, $819$ pairs were found.

\subsection*{Vulnerability analysis}

We investigated the vulnerability of various FRSs to morphs generated by our proposed architecture. Using a subset of bona fide images from the UNCW database and the morphed images generated by our architecture, we investigated the vulnerability of four different open-source FRSs and two COTS systems. While an open-source FRS can illustrate the applicability of the proposed approach, the evaluation of COTS systems will indicate a higher relevance for security considerations regarding Morphing Attacks in operational scenarios.

\subsubsection*{Calibration of the decision thresholds for verification}
\label{methods:calibration}

Since each of the open-source FRSs operates with a unique decision threshold, we first determined their respective thresholds specifically on the UNCW data set. We calibrated these thresholds for the four open-source FRSs which we used for face verification (which were the same as those used for pre-selection). To determine the respective thresholds, a subset of $500$ data subjects was sampled and all possible one-to-one combinations of mated pairs were obtained using each FRS. Similarly, all possible combinations of non-mated comparison scores were computed. As the total number of possible non-mated comparisons highly outnumbered the amount of possible mated comparison scores, a uniform sampling from all possible non-mated comparisons was performed to obtain an equal number.

Detection Error Trade-off (DET) curves were calculated for each FRS using the respective mated and non-mated distributions. The decision thresholds $\tau$ for False Match Rates (FMRs) of $0.1\%$ were empirically determined for each FRS \cite{manuel_aguado_martinez_pyeer_nodate} based on the FRONTEX recommendation \cite{frontex2015best}. 
The thresholds, along with the corresponding False Non-Match Rates (FNMRs), are shown in Table~\ref{table:thresholds}.

For face verification, we further deployed two COTS FRSs. For these, a default threshold was used to achieve an FMR of $0.1\%$ as recommended by the respective COTS vendors.

\begin{table}[!t] \centering
\caption{{\bf The verification thresholds on the unnormalized Cosine distances for each open-source FRS.} 
Thresholds were calculated on the UNCW data set. The corresponding FNMRs are illustrated next to the thresholds as decimal fractions.}
\label{table:thresholds}
\begin{tabular}{@{} c | c c @{}}    \hline
FRS & threshold @ FMR$=0.1\%$ & FNMR @ FMR$=0.1\%$  \\\hline
ArcFace   & 0.498 & 0.051 \\ 
DeepFace  & 0.125 & 0.784 \\ 
VGG-Face  & 0.146 & 0.318 \\ 
MagFace   & 0.666 & 0.004 \\ \hline

\end{tabular}

\end{table}

\subsubsection*{Vulnerability analysis metrics}

After determining the threshold, we analyzed the newly created morphed images for their attack potential using three different metrics, such as Product Average Mated Morph Presentation Match Rate (prodAvgMMPMR) (Eq~\ref{eq:prodAvgMMPMR}), Relative Morph Match Rate (RMMR) (Eq~\ref{eq:RIAPAR}), and Morphing Attack Potential (MAP). While prodAvgMMPMR and RMMR give the attack potential with respect to a single FRS, MAP gives the attack potential of the newly created data set across multiple FRSs. In this study, all rates are reported as decimal fractions and are therefore distributed in the interval $[0;1]$.

\paragraph*{MMPMR}
The Mated Morph Presentation Match Rate (MMPMR) \cite{scherhag_face_2016} is defined for distance scores (Eq~\ref{eq:MMPMR}),

    \begin{equation} \label{eq:MMPMR}
        MMPMR = \frac{1}{M} \sum_{m=1}^M \{  (  \min_{n=1,...,N_m} D_m^n ) < \tau \}
    \end{equation}

    with $M$ being total number of morphed images,
    $D_m^n$ the mated morph comparison score (here: distance score) of subject $n$ at morph $m$, $N_m$ the total number of subjects constituting to morph $m$, and $\tau$ the decision threshold.
    
\paragraph*{prodAvgMMPMR}

The Product Average Mated Morph Presentation Match Rate (prodAvgMMPMR) \cite{scherhag_biometric_2017} is a variant of MMPMR that allows a more probabilistic interpretation of the success of Morphing Attacks (Eq~\ref{eq:prodAvgMMPMR}),

    \begin{equation} \label{eq:prodAvgMMPMR}
        prodAvgMMPMR = \frac{1}{M} \sum_{m=1}^M [ \prod_{n=1}^{N_m} ( \frac{1}{I_m^n} \cdot \sum_{i=1}^{I_m^n} \{ D_m^{n,i} < \tau \} ) ]
    \end{equation}
    
    in which, additionally to the above,
    $I_m^n$ is the number of samples of subject $n$ within morph $m$, and $D_m^{n,i}$ the mated morph comparison score of sample $i$ of subject $n$ at morph $m$.
    
    An example: One morphed image was evaluated. Two data subjects contributed to the morph with one image each. Three bona fide samples per subject were tested against the morph. For one data subject, $\frac{2}{3}$ of the comparison scores exceeded the threshold $\tau$. For the other data subject, $\frac{3}{3}$ of comparison scores exceeded the threshold $\tau$. The prodAvgMMPMR then was simply the product of $\frac{2}{3}$ and $\frac{3}{3}$, therefore $\frac{2}{3}$.
    
\paragraph*{RMMR} 

    The Relative Morph Match Rate (RMMR) metric \cite{scherhag_biometric_2017} on the other hand takes the FNMR of a biometric system into account. Different biometric systems, calibrated at a particular FMR, can have different FNMRs. For instance, the FNMRs of the calibrated open-source FRSs greatly varied after calibration of the decision threshold (see Table~\ref{table:thresholds}).    
    If the FNMR is high, the system is less suited for an operation in a particular scenario, e.g., access control. Consequently, it might produce low MMPMR or prodAvgMMPMR -- therefore be less vulnerable to Morphing Attacks -- but at the same time rejects a large proportion of mated verification attempts. Therefore the RMMR relates the MMPMR to the FNMR (Eq~\ref{eq:RIAPAR}).
        
    \begin{equation}  \label{eq:RIAPAR}
        RMMR = MMPMR + FNMR 
    \end{equation}
    
    MMPMR and FNMR (and therefore RMMR) are specific for the chosen decision threshold $\tau$. Thus, if MMPMR is high, therefore the morphs would fool the FRS at $\tau$, and at the same time, if the FRS performs well by having a low FNMR, the RMMR would level off around $1$. On the other hand, if both the potential of the attack is low (low MMPMR), and the FRS also performs poorly by having a high FNMR, the RMMR would still level off at around $1$.  Most interestingly, if the potential of the attack is poor (i.e., low MMPMR), and the FRS performs well by having a low FNMR, the RMMR would be around $0$. For the sake of completeness: if the attack is of high quality (high MMPMR), and the FRS performs poorly (high FNMR), the RMMR could theoretically level off at $2$. However, that would require the morphed comparison distances to be smaller than the mated comparison distances.

\paragraph*{MAP}
\label{methods:mvr}
    
\begin{figure}[!t]
     \centering
     \includegraphics[width=1.\linewidth]{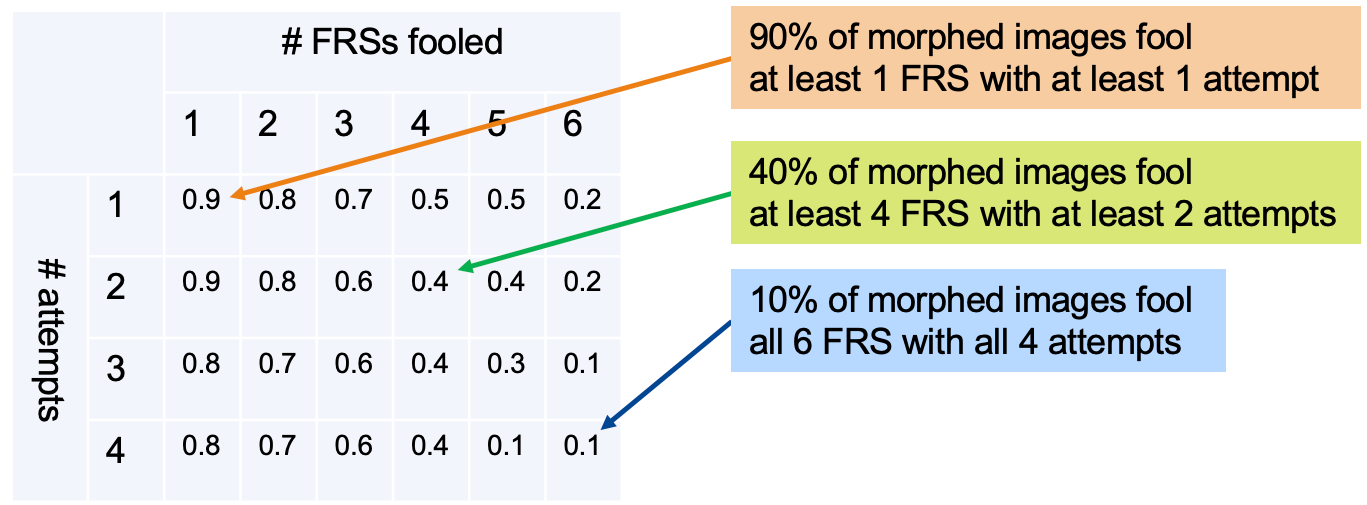}
     \caption{{\bf Morphing Attack Potential (MAP).}
     The MAP is a matrix describing the success of a data set of morphed images to fool a set of FRSs using multiple attack attempts. Several FRSs (x-axis) are attacked with several mated Morphing Attack attempts (y-axis). The element of a MAP matrix describes the proportion of successful verifications of both attackers (i.e., both contributing subjects of each morph) at a given number of attempts (i.e.,  number of different bona fide images for both subjects) and with a particular number of fooled FRSs. Note that MAP was calculated as a decimal fraction within the range $[0;1]$.}
     \label{fig:mvr_explanation}
    \end{figure}

    Recently, the Morphing Attack Potential (MAP) has been proposed to report the attack potential of a data set $D$ of morphed images in a combined manner across different FRSs \cite{ferraraMorphVulnerabilityRate2022, isoiec_international_2023}. All FRSs (in our case $6$ different systems) verified the same number of different bona fide images (e.g. $4$) of each subject against the respective morph. $MAP_{4,6}^D$ then represents the $4\times6$ matrix, where the element $(i,j)$ indicates the decimal fraction of morphed images for which at least $i$ verification attempts were successful with respect to both contributing subjects and at least $j$ FRSs (Fig~\ref{fig:mvr_explanation}). As outlined earlier, MAP values are characterized by higher generalizability and robustness compared to many other metrics (cf. Introduction). The MAP is now adopted in the ISO/IEC CD2 20059 standard \cite{isoiec_international_2023}. The standard further suggests the MAPavg metric, a weighted average which reduces the MAP matrix to one scalar. The weights are set higher for the cells to the lower and to the right of the MAP \cite{isoiec_international_2023}. We averaged the MAPs for different morphing algorithms before calculating one MAPavg per pre-selection method. Additionally, we evaluated the attack potential using the recently proposed G-MAP metric, which similarly aims at reducing MAP to one single scalar \cite{singh_deep_2023}. We have calculated G-MAP for each pre-selection method separately, across morphing algorithms and verification systems. The calculations of MAPavg and G-MAP differ in several aspects, e.g., weighting is performed in MAPavg, but a different number of attacks per morph is allowed in G-MAP.
    
\subsection*{Morphing Attack Detection} 
\label{methods:MAD}

\begin{figure}[!t]
	\centering
	\includegraphics[width=1.\linewidth]{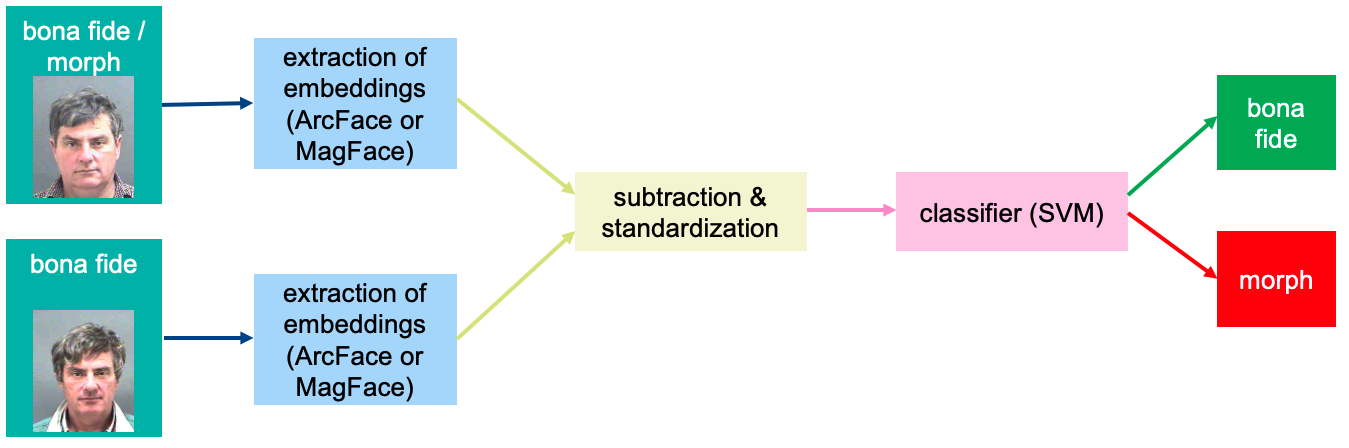}
	\caption{{\bf D-MAD pipeline.}
    ArcFace or MagFace embeddings were extracted from bona fide images and morphed images. Differential embeddings have been created by subtraction of either the embeddings of a bona fide image from a morphed image or by the subtraction of a bona fide image from a different bona fide image of the same data subject. The differential vectors have been re-scaled to $N(0,1)$. A classifier was trained (on ArcFace and MagFace differential embeddings, separately) to differentiate between bona fide images and morphed images.\\ \emph{Face images are republished from \cite{uncw_morph_nodate} under a CC BY license, with permission from Prof. Karl Ricanek Jr, University of North Carolina at Wilmington, original copyright 2006.}}
	\label{fig:mad_pipeline}
\end{figure}

In the previous parts, we described the methodology for morphing and evaluating the vulnerability of FRSs. In addition, the study at hand uses embeddings to detect Morphing Attacks.
In particular, we draw inspiration from Differential Image Morphing Attack Detection (D-MAD) approaches, which compare a presented image with a trusted bona fide image to evaluate the nature of the presented image. We used a D-MAD approach proposed by Scherhag et al. \cite{scherhag_deep_2020} as our baseline. The chosen D-MAD approach performs a differential analysis of ArcFace embedding vectors to train a binary Support Vector Machine (SVM) classifier using radial basis functions and else default parameters as implemented in sklearn (v. 0.24.2). Specifically, the ArcFace embeddings were extracted from the suspicious images to be analyzed. ArcFace embeddings were further extracted from bona fide probe images of one of the participating morph candidates. These bona fide images are comparable to trusted live captures of an attacker. The procedure is outlined in Fig~\ref{fig:mad_pipeline}. The two embedding vectors were then subtracted from each other. The resulting difference vectors of length $512$ portray the samples of morphed (differential) images. As samples of bone fide (differential) images, the same procedure has been carried out by subtracting the embeddings of two different bona fide captures of the same data subject. The resulting difference vectors were scaled to follow a standard Normal distribution with $\mu=0$ and $\sigma=1$ which were then handed over as features to the SVM.

While we found a decent performance of the previously proposed D-MAD approach, we want to note that the embeddings of MagFace instead of ArcFace could raise the recognition accuracy to a new level. The loss function of MagFace is designed in such a way that it not only arranges the samples of a class (a subject) adjacent in the multidimensional space. It is further designed so that samples with higher quality, or samples for which the certainty of class membership is high, are closer to the center of the class \cite{meng_magface_2021}. Thus, the distances in the embedding space between two samples of the same class, which are of high quality, or, conversely, are certain to belong together, are very small. 
On the other hand, the distance in the multidimensional space between two samples is quite large when the membership estimate of one of the samples is less accurate due to low image quality. The size of MagFaces' embedding vector increases monotonically with image quality. This results in a larger difference between the two embedding vectors when the quality of a face image is low. Using MagFace instead of ArcFace embeddings could not only lead to the reported superior performance of MagFace in face recognition. It could also combine the strengths of an embedding-based D-MAD approach such as that of Scherhag et al. \cite{scherhag_deep_2020} with approaches based on image quality analysis, such as the approach of Venkatesh et al. \cite{venkatesh_detecting_2020}.

In the present study, the procedure closely followed the approach described by Scherhag et al. \cite{scherhag_deep_2020} using ArcFace embeddings. However, the same approach has then been repeated in an analogous fashion using MagFace embeddings (Fig~\ref{fig:mad_pipeline}).

\subsubsection*{Training protocol}
Only a subset of $80\%$ of the generated control data set was used for training. This control data set consisted of morphed images without pre-selection based on embeddings, but face images were randomly morphed after demographic consistency checks (see above). For training, morphed images from all morphing algorithms were used together. Thus, from a pair of faces selected (randomly) for morphing, the morphs of all four morphing algorithms used were placed in either the training or the test set. A subset of $80\%$ of all non-morphed subjects (with at least $2$ face samples, which was about $10,000$ subjects) was used for training on the bona fide differential embeddings, and correspondingly $20\%$ for testing. More importantly, testing was also performed on all morphs that were generated based on pre-selection using our proposed architecture, namely using distances between embeddings from different FRSs, such as ArcFace, DeepFace, VGG-Face, \& MagFace.

\subsubsection*{Testing metrics}
\label{methods:abpcer}

To evaluate the MAD algorithms, ISO/IEC 30107-3 and ISO/IEC CD2 20059 \cite{isoiec_international_2017, isoiec_international_2023} propose to calculate the Morph Attack Classification Error Rate (MACER) and the Bona fide Presentation Classification Error Rate (BPCER). MACER was formerly named APCER (Attack Presentation Classification Error Rate) but renamed in the context of Morphing Attacks \cite{isoiec_international_2023}. Similar to the metrics used in the previous analyses, all rates will be reported as decimal fraction in a range of $[0;1]$.

MACER subserves as a security measure, i.e., the proportion of attack presentations incorrectly classified as bona fide presentations must be small for a secure biometric system. On the contrary, BPCER subserves as a convenience measure, i.e. a low number of false negatives is aimed for in an operational biometric system.
Oftentimes, the BPCER10 is also reported \cite{scherhag_deep_2020}. BPCER10 is the BPCER at the threshold of the system, at which the MACER is $10\%$, i.e., $0.1$ \cite{scherhag_deep_2020}. BPCER10 can subserve as a convenience metric, at a given security level.

\section*{Results}

\subsection*{Vulnerability analysis of FRSs}
\label{results:vulnerability}

\subsubsection*{Mated morph comparison performance}

\begin{figure}[!h]
 \centering
 \includegraphics[width=1.\textwidth, trim={0 0 0 0}]{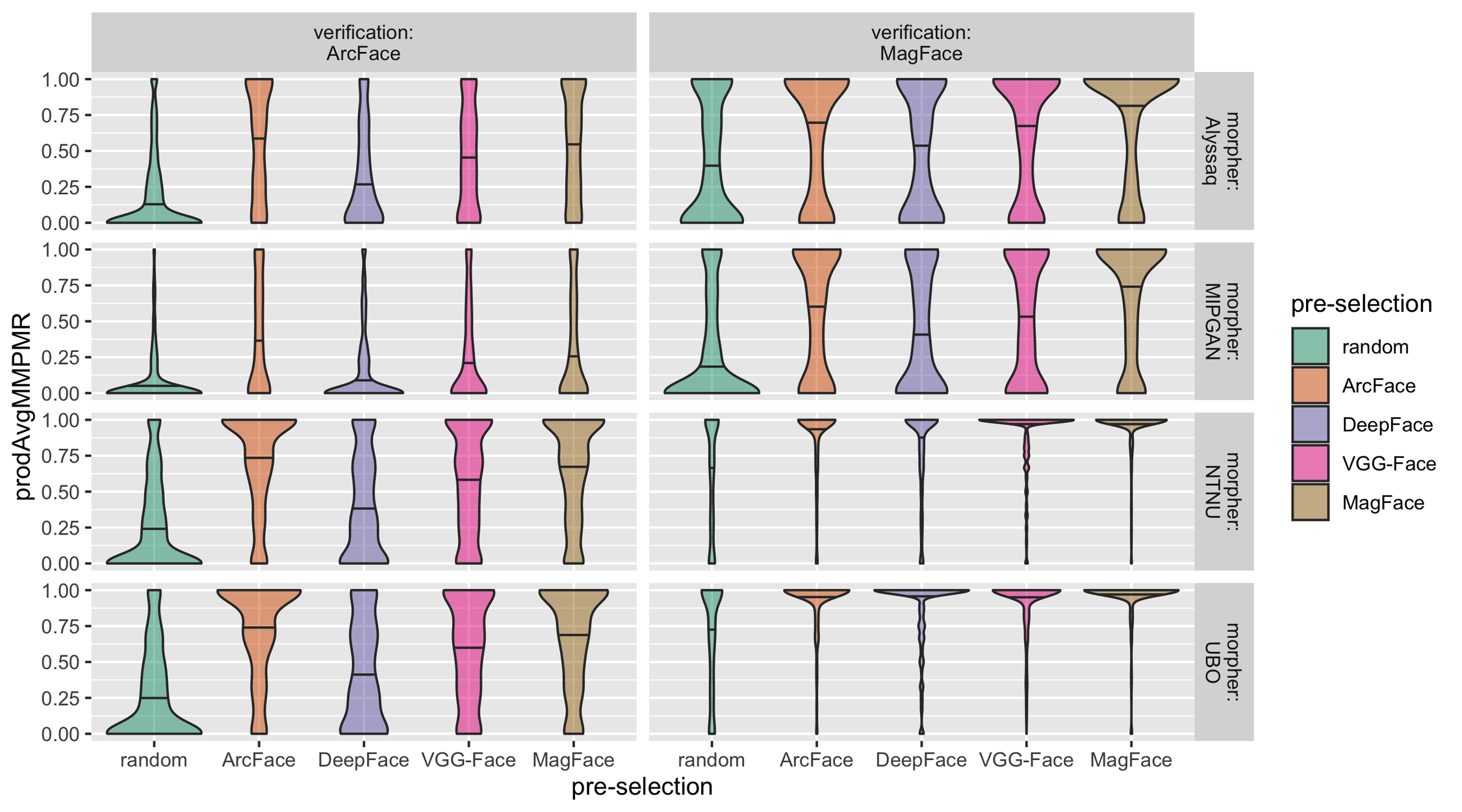} 
 \caption{{\bf Mated morphs comparison success rates for different image pre-selection embeddings.} 
 prodAvgMMPMRs (y-axes) are plotted for different pre-selection methods (x-axis \& color-coded). Density is plotted in horizontal direction. Median values are illustrated by horizontal black bars. The same pairs were morphed by different morphing methods (rows). Random assignments of the morphing pairs are displayed in the left-most column. All morphs were verified using ArcFace and MagFace (columns). See \ref{fig:res:MMPMR_MORPH_DF_VF} for verifications using DeepFace and VGG-Face. Note that prodAvgMMPMR was calculated as a decimal fraction within the range $[0;1]$.}

 \label{fig:res:MMPMR_MORPH_AF_MF}
\end{figure}

We studied the vulnerability of different FRSs when attacked by the data set we created. Specifically, we selected two different FRSs, namely ArcFace \cite{deng_arcface_2019} and MagFace \cite{meng_magface_2021}, to illustrate their vulnerability to face images generated by our proposed architecture. The corresponding success rates were measured using prodAvgMMPMR and are shown in Fig~\ref{fig:res:MMPMR_MORPH_AF_MF}. Moreover, the proposed approach is investigated using four different morphing algorithms.

As shown in Fig~\ref{fig:res:MMPMR_MORPH_AF_MF}, image pre-selection increased attack potential as compared to random pairing  when the resulting morphs were verified using ArcFace or MagFace. The attack potential increased when MagFace was used as the verification system followed by ArcFace (Fig~\ref{fig:res:MMPMR_MORPH_AF_MF}). While we also evaluated two other FRSs based on VGG-Face and DeepFace, the attack potential did not increase as the FRSs by themselves were relatively low performing (\ref{fig:res:MMPMR_MORPH_DF_VF}), Table~\ref{table:thresholds}. See below for a detailed analysis of this behavior.

We further note a link between the FRS used for pre-selection and the FRS used for the assessment of vulnerability. If the pre-selection was based on the same FRS which is also used to assess vulnerability, the attack potential of the database was higher. This does not come as a surprise, since the embeddings are treated the same in both cases. However, when COTS FRSs were used, the attack potential was still increased compared to random pairing, although it was not biased by using the same FRSs twice within the analysis pipeline. The vulnerability to the morph attacks for the two COTS FRSs tested is illustrated in \ref{fig:res:MMPMR_MORPH_COTS}. For the COTS FRSs, the prodAvgMMPMR mostly accumulated around $1$, indicating an extremely high vulnerability even for morphs based on random pre-selection (\ref{fig:res:MMPMR_MORPH_COTS}). In addition, the morphs created with MIPGAN and verified with COTS FRSs again illustrate the benefit of image pre-selection (\ref{fig:res:MMPMR_MORPH_COTS}).

Importantly, the increased attack potential, regardless of the morphing algorithm used, can be clearly observed throughout the proposed pre-selection approach. However, there was a noticeable difference in the success of Morphing Attacks. NTNU morpher and UBO morpher produced the best Morphing Attacks, followed by Alyssaq morpher and lastly MIPGAN (Fig~\ref{fig:res:MMPMR_MORPH_AF_MF} \& \ref{fig:res:MMPMR_MORPH_COTS}).

The MAP has recently been introduced as a general measure of the success of Morphing Attacks across different verifying FRSs \cite{isoiec_international_2023, ferraraMorphVulnerabilityRate2022}. Briefly, the elements of a MAP matrix contain the proportions of successful Morphing Attacks (with both data subjects involved) that fool a given number of FRSs for a given number of attack attempts (Fig~\ref{fig:mvr_explanation}). The higher the values, and the further the high values spread to the lower right of the matrix, the more effective the attacks were on the tested data set.

\begin{figure}[!h] 
 \centering
 \includegraphics[width=0.9\textwidth, trim={0 0 0 0}]{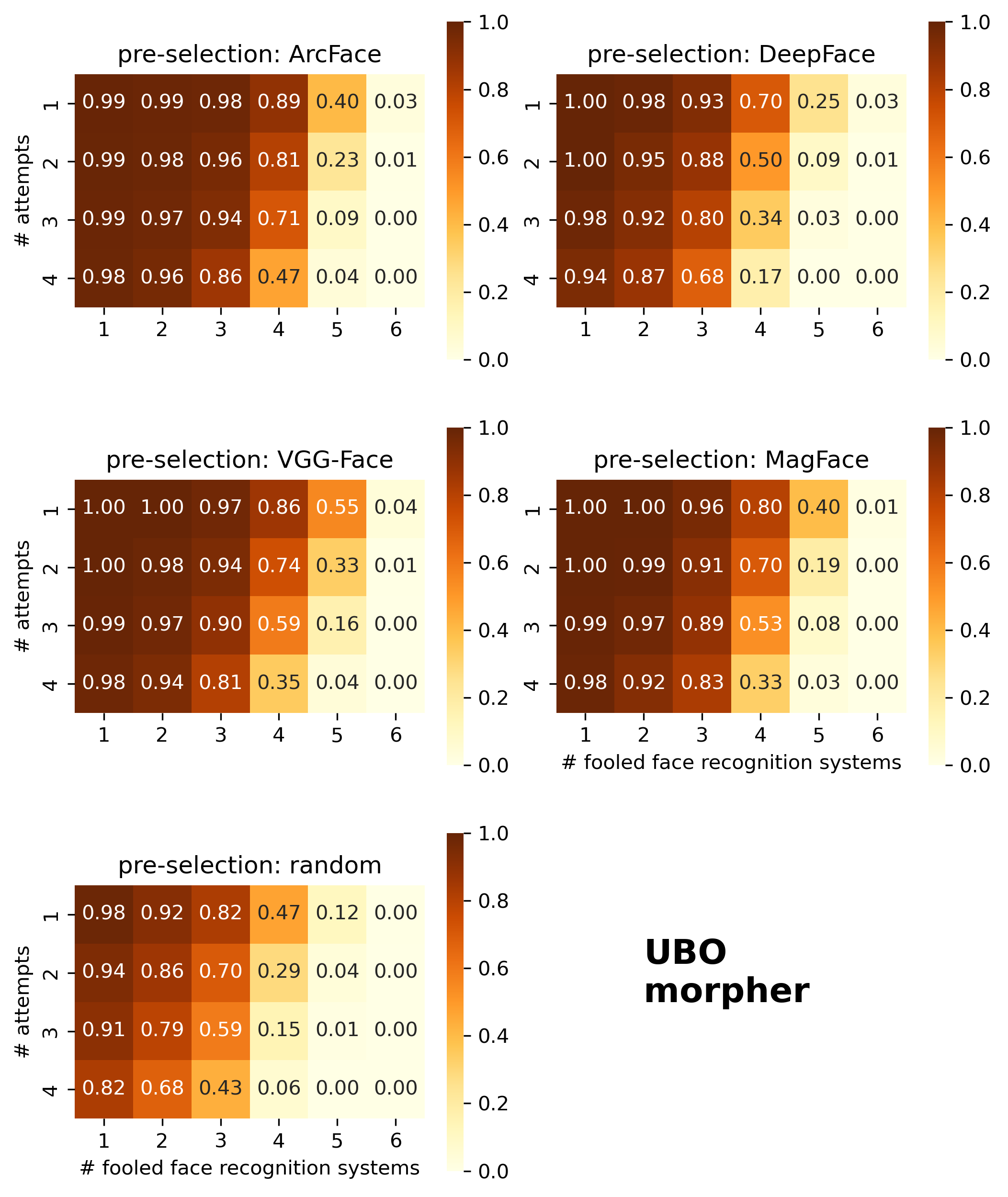} 
 \caption{{\bf Morphing Attack Potential (MAP) of morphs generated by the UBO morpher.}
 Different FRSs were used for image pre-selection, i.e. ArcFace, DeepFace, VGG-Face, or MagFace (different heatmaps). Alternatively, pairs were randomly assigned (bottom heatmap). For each FRS used for pre-selection, the resulting morphs were verified against four bona fide images of each subject. The ratio of successful attempts for both subjects is shown on each y-axis of each plot. In addition, different FRSs were used to verify the paired morphs, four open-source FRSs and two COTS FRSs. The percentage of successful attacks across multiple FRSs is plotted on each x-axis. The MAP is shown and color-coded in each cell and describes the proportion of successful verifications for a given number of attempts (y-axes) and FRSs (x-axes). Note that the MAP was calculated as a decimal fraction in the range $[0;1]$.}
 \label{fig:MVR_UBO}
\end{figure}

Fig~\ref{fig:MVR_UBO} shows MAPs for morphs created by the UBO morpher. Again, using pre-selection generally increased the MAPs. All non-random pre-selection methods successfully outwitted at least four (out of six) FRSs in about $70$ to $90\%$ of cases with at least one attack attempt. In contrast, random morphs only exceeded $47\%$. In about $17\%$ to $47\%$ of cases, all four attack attempts were able to fool four different FRSs when pre-selection was performed. However, only single-digit percentages of morphs were able to fool four FRSs with all four attack attempts.

MAPs were comparable when morphs were created by the NTNU morpher instead of the UBO morpher (\ref{fig:MVR_NTNU}). However, MAPs were significantly lower when morphs were created by the Alyssaq morpher (\ref{fig:MVR_ALYSSAQ}), and even lower for morphs created by MIPGAN (\ref{fig:MVR_MIPGAN}). However, a definite distinction between morphs pre-selected by different FRSs embeddings is more complex. 

Therefore, we further calculated MAPavg for each pre-selection method \cite{isoiec_international_2023}, which aggregates the MAP matrix into a single scalar by performing a weighted average, and with higher values corresponding to stronger attack potentials. The weights were defined by the positions of the cells, i.e., morphs with more successful attack attempts were weighted higher, as well as attacks which fool more FRSs. Values were calculated as decimal fraction, with one MAPavg value per pre-selection method. When verifying with six FRSs (four open-source and two COTS), MAPavg were $0.173$ (no pre-selection), $0.247$ (DeepFace), $0.322$ (VGG-Face), $0.339$ (ArcFace), and $0.317$ (MagFace). Additionally, we removed DeepFace and VGG-Face from the verifying FRSs of the MAPavg analyses, as they showed high FNMRs and would therefore not be used in real-world scenarios (see next Section for more details). The resulting MAPavg were $0.347$ (no pre-selection), $0.487$ (DeepFace), $0.580$ (VGG-Face), $0.662$ (ArcFace), and $0.614$ (MagFace).

Similarly, we calculated G-MAP for each pre-selection method \cite{singh_deep_2023}, with higher values corresponding to stronger attack potentials. When calculating G-MAP across all morphing algorithms and six FRSs for verification, the resulting G-MAP values for each pre-selection FRS were $0.358$ (no pre-selection), $0.465$ (DeepFace), $0.549$ (VGG-Face), $0.558$ (ArcFace), and $0.564$ (MagFace). By also removing DeepFace and VGG-Face as verifying systems, the G-MAP values for each pre-selection method were $0.522$ (no pre-selection), $0.667$ (DeepFace), $0.746$ (VGG-Face), $0.793$ (ArcFace), and $0.809$ (MagFace). 

The results confirm that pre-selection was better than none, and MagFace and ArcFace generated the strongest attacks (depending on the metric used for evaluation), followed by VGG-Face, and lastly DeepFace.

\subsubsection*{Relative mated morph comparison performance}
\label{sec:chapter03:riapar}

\begin{figure}[p]
 \centering
 \includegraphics[width=1.\textwidth, trim={0 0 0 0}]{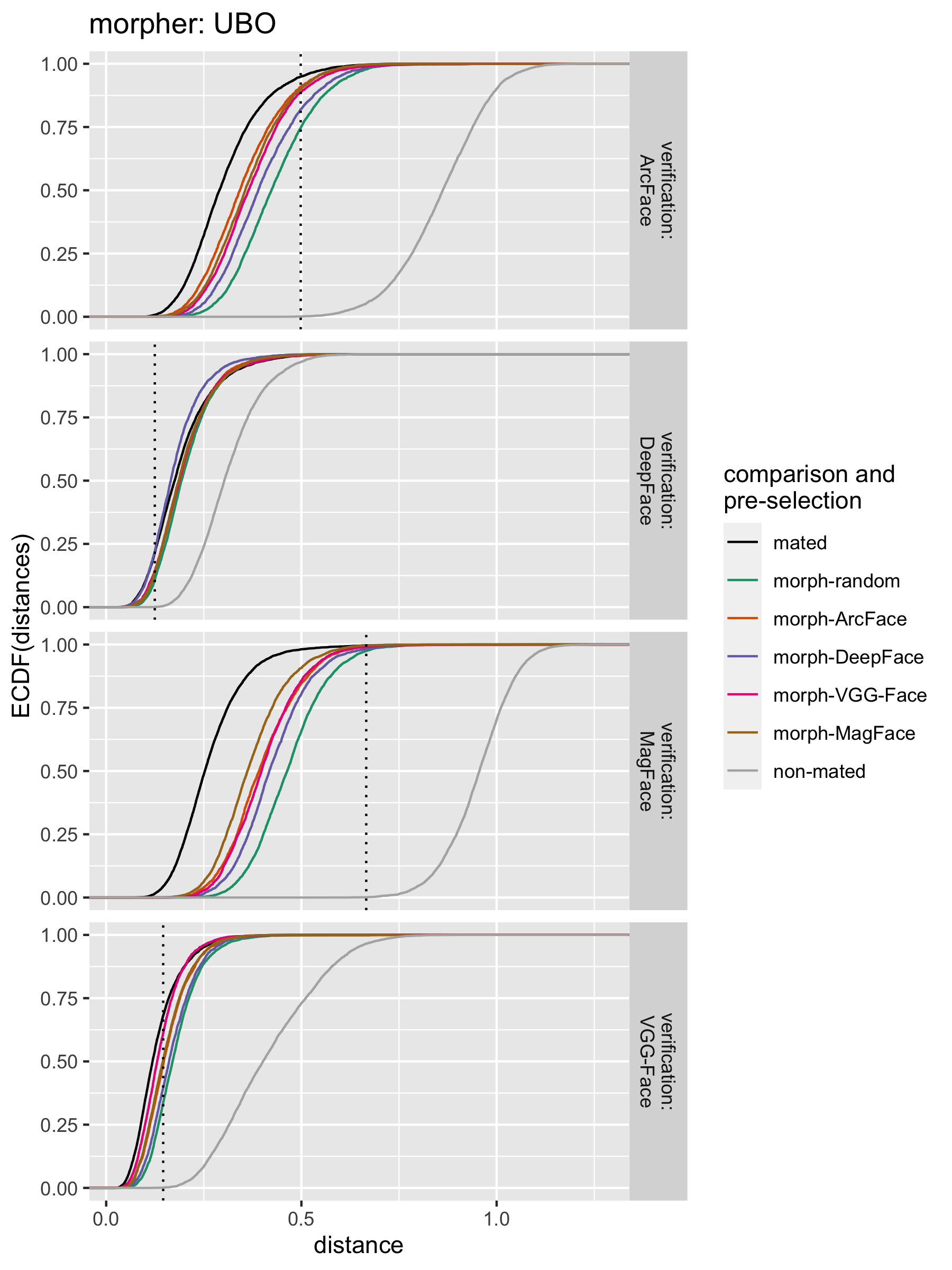} 
 \caption{{\bf ECDFs for distance scores of the open-source FRSs.}
 Mated, non-mated, and mated morph comparisons were performed. Morphs were created using the UBO morpher. The distance values for the comparisons are shown on the x-axis. The (cumulative) proportion of positive verifications at a certain distance score is plotted on the y-axes. Different FRSs were used for verification (rows). The different types of comparisons are color-coded, i.e., mated, non-mated, or mated morph comparisons, including morphs pre-selected with the help of face embeddings of the different FRSs. The dotted vertical lines indicate the $0.1\%$ FMR threshold for each FRS used for verification.}
 \label{fig:ecdf_ubo}
\end{figure}

To further examine the performances of data sets using the different pre-selection methods, as well as the behavior of the verification algorithms, the distributions of the raw distance scores of mated comparisons, non-mated comparisons, and mated morph comparisons were visualized as Empirical Cumulative Distribution Functions (ECDFs) in Fig~\ref{fig:ecdf_ubo}, using morphs created with the UBO morpher as an example. 
Across all four open-source verification FRSs, the mated morph comparison scores were distributed between mated scores and non-mated scores. However, they were closer aligned to the mated scores than to the non-mated scores, even for morph pairs without pre-selection (i.e., random assignment). Importantly, morphs pre-selected with our proposed architecture performed better than morphs from random pre-selection. Similarly to before, the same verification FRS was biased for morphs pre-selected by their own embeddings prior to morphing.

The comparison decision highly varied between the verification FRSs. Whereas DeepFace incorrectly accepted only a very small number of morphs, followed by VGG-Face, ArcFace, and most significantly MagFace incorrectly accepted nearly all morphs as mated comparisons. On the contrary, at the calibrated threshold of FMR $=0.1\%$, DeepFace and to a lesser extent VGG-Face both revealed high FNMRs (Table~\ref{table:thresholds}), therefore incorrectly rejecting a large proportion of mated verification attempts. On the opposite, ArcFace, and more importantly, MagFace had very low FNMRs at the given FMR (Table~\ref{table:thresholds}). This has led to a higher susceptibility of \emph{better} FRSs -- in the sense of low FNMR at a given FMR -- to Morphing Attacks.

We call this phenomenon \emph{morphing attack paradox}. The better the FRS and therefore the lower the FNMR of the FRS on a preset threshold, the more tolerant the FRS is to mated presentations. The more tolerance the FRS shows to mated presentations, the more susceptibility it is to Morphing Attacks. As a result, more accurate FRSs are more susceptible to Morphing Attacks.

\ref{fig:ecdf_ntnu}, \ref{fig:ecdf_alyssaq}, \& \ref{fig:ecdf_mipgan} illustrate the respective distributions of distance values for morphs created with the other morphing algorithms. The general patterns were the same as in Fig~\ref{fig:ecdf_ubo}. However, while the distance distributions of mated morph comparisons with NTNU morphs closely resembled those of morphs created with the UBO morpher, morphs created with Alyssaq and MIPGAN showed higher relative distances, resulting in a higher number of rejections of mated morphs at the given decision thresholds.

Since mated morph distances of more accurate FRSs -- such as ArcFace and MagFace -- were distributed between mated distances and non-mated distances, Fig~\ref{fig:ecdf_ubo} indicates that there is a chance of separating mated morph comparisons from mated comparisons by adjusting the decision threshold of the FRS. Such an adjustment could dramatically reduce the vulnerability for MagFace, for which the distance distributions of mated comparisons and mated morph comparisons showed only a slight overlap. Using ArcFace for verification, the overlap between distributions was already stronger. Therefore, threshold adjustment for verification would lead to significantly higher FNMRs in ArcFace. Contrarily, the distributions of mated morph distances of less accurate FRSs such as DeepFace and VGG-Face closely aligned to the distribution of the mated distances (Fig~\ref{fig:ecdf_ubo}). In the case of DeepFace, especially when both image pre-selection and verification were performed with the same FRS, the mated morph distances were even smaller than the mated distances.

\begin{figure}[!t]
 \centering
 \includegraphics[width=1.\linewidth, trim={0 0 0 0}]{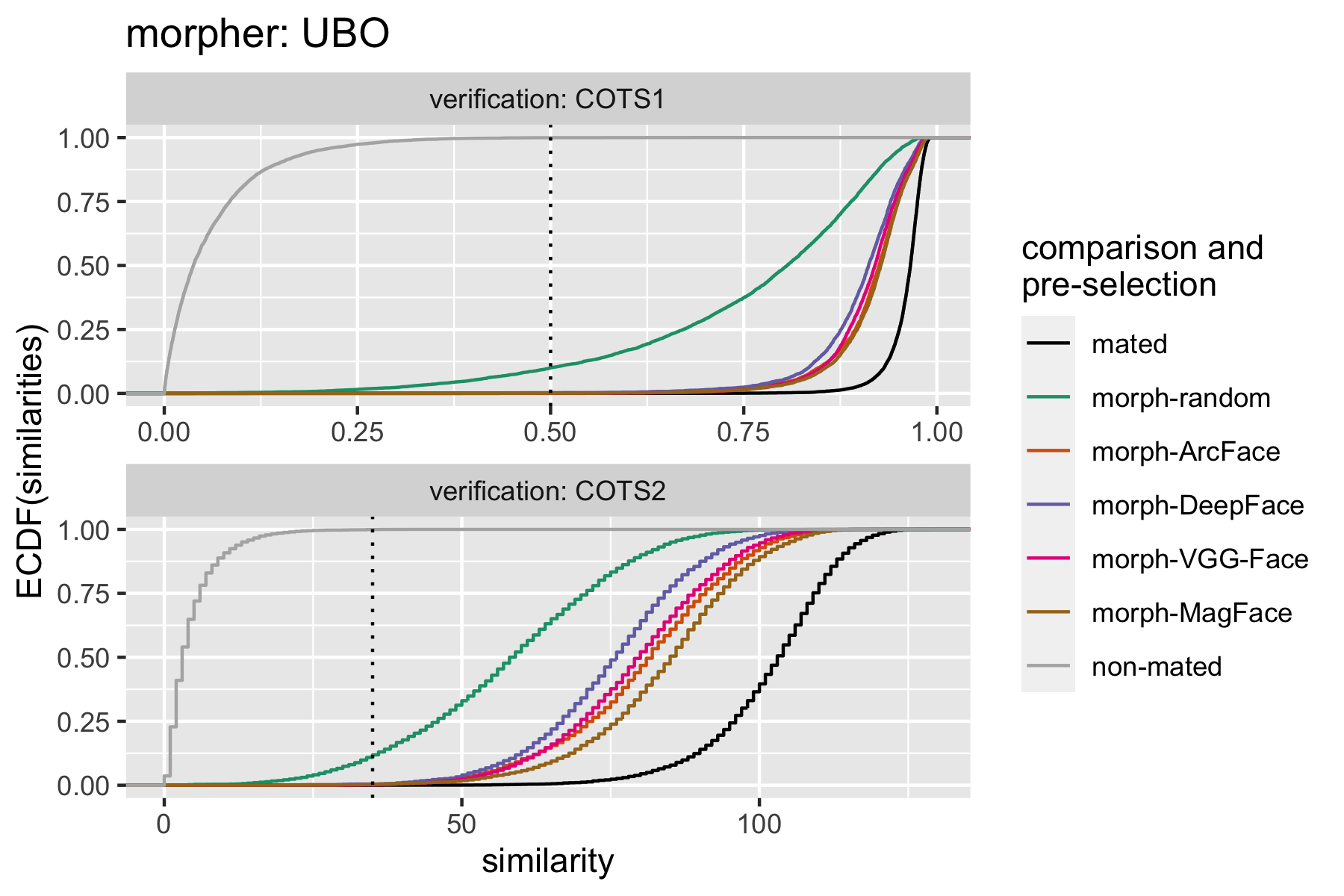} 
 \caption{{\bf ECDFs for similarity scores of the COTS FRSs.}
 Mated, non-mated, and mated morph comparisons were performed. Morphs were generated using the UBO morpher. The different similarity scores for the comparisons are displayed on the x-axis. The (cumulative) proportion of successful verifications at a particular similarity score is plotted at the y-axes. Note that because similarities instead of distances were used, the interpretation of the x-axes must be flipped compared to Fig~\ref{fig:ecdf_ubo}. Different COTS FRSs were used for verification (rows). The different types of comparisons are color-coded, i.e., mated, non-mated, or mated morph comparisons, with morphs pre-selected with the help of face embeddings of certain FRSs. The dotted vertical lines indicate the $0.1\%$ FMR threshold for each FRS used for verification.}
 \label{fig:ecdf_ubo_cots}
\end{figure}

Fig~\ref{fig:ecdf_ubo_cots} further illustrates the ECDFs of the similarity scores using COTS FRSs and the UBO morpher (see \ref{fig:ecdf_ntnu_cots}, \ref{fig:ecdf_alyssaq_cots}, \& \ref{fig:ecdf_mipgan_cots} for the morphs created by the other morphers). Since the COTS FRSs were \emph{not} used for pre-selection, the results are less biased with respect to the pre-selection algorithm. First, even with random pre-selection, all types of morphs were likely to be successfully verified by the COTS FRSs. However, similar to the open-source FRSs, the distributions of the mated morph comparisons shifted toward the distributions of the mated comparisons, when pre-selection according to our architecture was applied. A hierarchy between the different pre-selection methods can be seen. Morphs derived from a pre-selection approach using MagFace embeddings produced the highest similarity scores, followed by ArcFace, VGG-Face, and finally DeepFace.

To further account for the performance of the individual FRSs, the RMMR was calculated using the open-source FRSs for verification. The RMMR corrects the MMPMR for the FNMR (Eq~\ref{eq:RIAPAR}). Thus, the strong inflation of the mated morph comparison values of the previous chapter can be corrected, especially for FRSs with high FNMRs. Table~\ref{table:riapar} shows the RMMR values for differently pre-selected, morphed, and verified images. A similar pattern as before can be seen. When the same FRS is used for pre-selection and verification, the RMMR is highest in most cases. However, the second highest RMMR is often obtained from a pre-selection with MagFace, followed by ArcFace and VGG-Face. Higher RMMR can also be observed for morphs created by the UBO morpher and the NTNU morpher compared to other morphers.  

\begin{table}[!t] \centering
\caption{{\bf Relative Morph Match Rates (RMMRs).} 
Images were morphed using different morphing algorithms, pre-selected using embeddings of different FRSs or alternatively, randomly pre-selected, and verified using different FRSs. The RMMR corrects the MMPMR by the FNMR of the verification FRS (see Eq~\ref{eq:RIAPAR}). The highest values row-wise are highlighted in bold, leaving out the quasi-diagonal elements, i.e., if pre-selection and verification FRSs coincided. Note that RMMR was calculated as a decimal fraction within the range $[0;1]$.}
\label{table:riapar}
\begin{tabular}{@{} c | c | c c c c c @{}}    \hline
verification & morpher & \multicolumn{5}{c}{pre-selection  } \\
 &  & random & ArcFace & DeepFace & VGG-Face & MagFace \\ 
 \hline
ArcFace  & Alyssaq & 0.24 &0.64 &0.39 & 0.55 &\textbf{0.63} \\
DeepFace &         & 0.78 &0.78 &0.78 & 0.78 &0.78\\
VGG-Face &          & 0.35 &0.44 &0.37 & 0.60  &\textbf{0.45} \\
MagFace  &          & 0.44 &0.64 &0.52 &\textbf{0.65} &0.76 \\\hline

ArcFace & UBO & 0.32 & 0.79 & 0.51 & 0.65 & \textbf{0.72} \\
DeepFace & & 0.79 & \textbf{0.81} & 0.84 & 0.80 & \textbf{0.81} \\
VGG-Face & & 0.42 & 0.57 & 0.45 & 0.71 & \textbf{0.58} \\
MagFace  &  & 0.72 & \textbf{0.95} & 0.87 & \textbf{0.95} & 0.97 \\\hline

ArcFace  &NTNU & 0.31 & 0.78 & 0.49 & 0.64 & \textbf{0.72} \\
DeepFace &          & 0.79 & 0.80 & 0.84 & \textbf{0.81} & \textbf{0.81} \\
VGG-Face & & 0.40 & 0.53 & 0.45 & 0.71 & \textbf{0.55} \\
MagFace  &  & 0.65 & \textbf{0.91} & 0.83 & \textbf{0.91} & 0.97 \\\hline

ArcFace  &MIPGAN & 0.13 & 0.44 & 0.22 & 0.32 & \textbf{0.37} \\
DeepFace &         & 0.79 & 0.79 & 0.80 & 0.79 & \textbf{0.80} \\
VGG-Face & & 0.33 & \textbf{0.37} & 0.35 & 0.44 & \textbf{0.37} \\
MagFace  &  & 0.28 & \textbf{0.60} &  0.45 & 0.54 & 0.68 \\\hline

 \hline
\end{tabular}

\end{table}

Table~\ref{table:riapar} can be summarized in the following fashion. To get some idea about how well the individual pre-selection FRSs have performed across morphing algorithms and open-source verification FRSs -- using RMMR as a metric -- each row of Table~\ref{table:riapar} was converted to ranks ($1$ to $5$). $5$ indicated the FRS for pre-selection (columns) that had the highest RMMR value compared to the other elements, and $1$ indicated the FRS with the lowest RMMR value, respectively. If equal values coincided in a row, decimal numbers were used. The ranks were then averaged across rows, therefore averaged across morphing algorithms and the attacked FRSs. Table~\ref{table:RMMR_ranks} illustrates the average ranks for the different pre-selection methods. Pairs based on MagFace embeddings generated the highest RMMR values, followed by ArcFace, VGG-Face, and finally, DeepFace. Randomly pre-selected pairs performed the worst across different morphing algorithms and verification systems.

\begin{table}[!ht] \centering
\caption{{\bf Average ranks for RMMR values for the different pre-selection methods.}
Pre-selection was either performed using random assignment of pairs or based on embeddings of four different FRSs.}
\label{table:RMMR_ranks}
\begin{tabular}{@{} c c @{}}  \hline  
 pre-selection  & average rank \\ \hline
        random & 1.13 \\
       ArcFace & 3.63 \\
      DeepFace & 2.63 \\
      VGG-Face & 3.56 \\
       MagFace & 4.06 \\
 \hline
\end{tabular}

\end{table}

\subsection*{Morphing Attack Detection performance} 
\label{results:MAD}

\begin{figure}[!b]
 \centering
 \includegraphics[width=.8\linewidth, trim={0 0 0 0}]{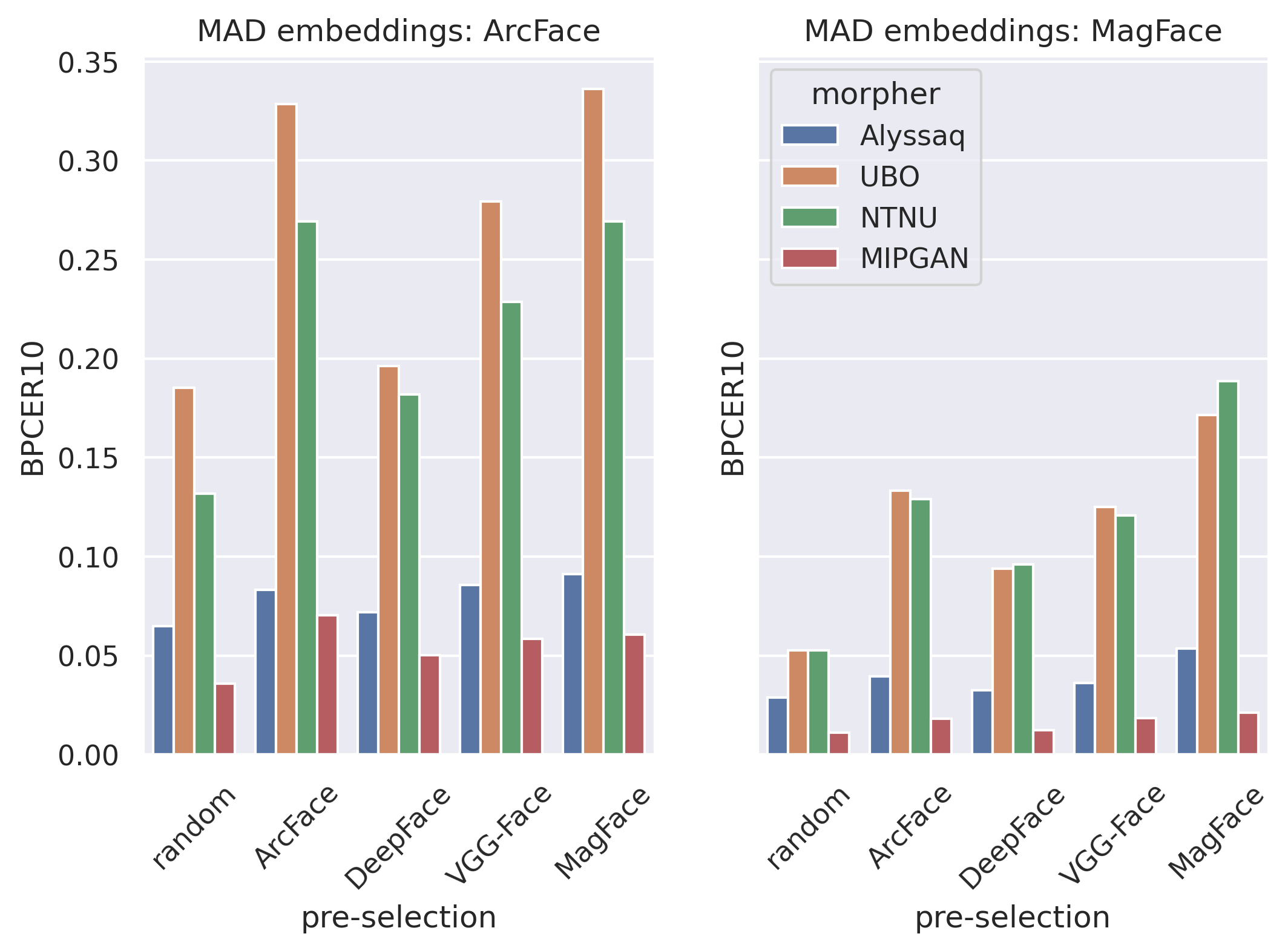} 
 \caption{{\bf D-MAD algorithm performances.}
 BPCER10 values of the classifiers tested on differently morphed and differently pre-selected data sets are shown. Left: metrics from a D-MAD algorithm trained on ArcFace embeddings. Right: Metrics from a D-MAD algorithm trained on MagFace embeddings. The images morphed by different morphing algorithms are shown in different colors. The pre-selection methods to generate the pairs for morphing are distributed along the x-axes. Note that BPCER10 was calculated as a decimal fraction within the range $[0;1]$.}
 \label{fig:mad}
\end{figure}

Fig~\ref{fig:mad} shows the corresponding BPCER10 values of the MAD classifiers, tested on morphs with different pre-selection applied, and corresponding bona fide images. The operational point values BPCER10 were lower for the D-MAD classifier trained with MagFace differential embeddings than for the one trained with ArcFace differential embeddings. Furthermore, the BPCER10 values on the test data sets were the lowest for randomly pre-selected pairs for morphing. The BPCER10 values were higher when the test set contained morphs of pairs that were pre-selected according to our proposed architecture. This trend was more pronounced for morphs generated by the NTNU morpher and even more in morphs generated by the UBO morpher. On the other hand, morphs generated by the Alyssaq morpher or MIPGAN did not lead to a pronounced increase in BPCER10 values.

\begin{figure}[!hb]
 \centering
 \includegraphics[width=.7\textwidth, trim={0 0 0 0}]{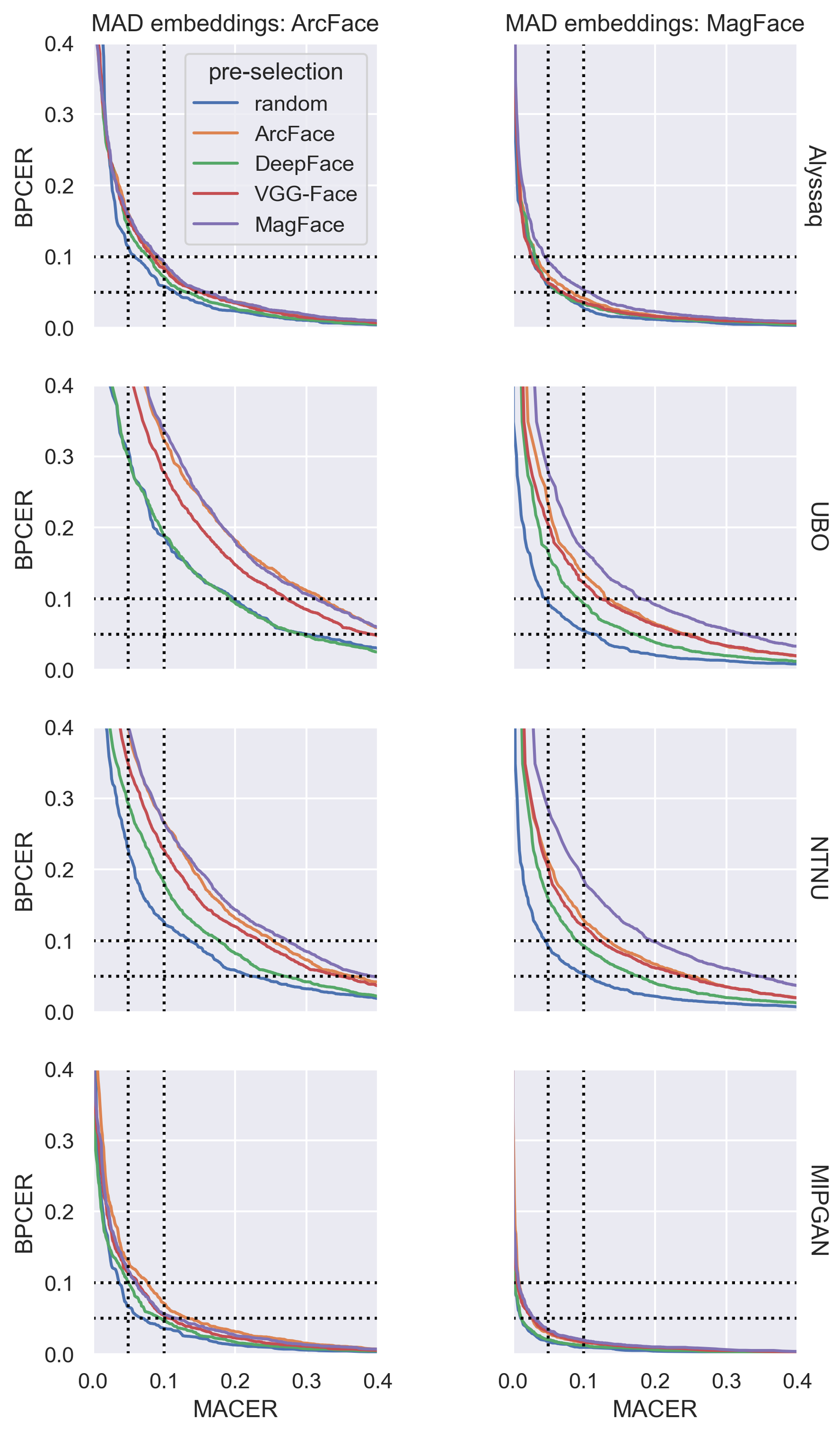} 
 \caption{{\bf DET curves of the D-MAD approaches.}
 Left column: D-MAD approach using ArcFace embeddings (original version). Right column: D-MAD approach using MagFace embeddings. Morphs of the different morphing algorithms are separated by rows. Data subsets of differently pre-selected morph pairs are color-coded. The BPCER is plotted against the MACER. Dotted lines indicate the positions where BPCER or MACER are $0.1$ (i.e., $10\%$) and $0.05$ (i.e., $5\%$). Note that both rates were calculated as decimal fractions within the range $[0;1]$.}
 \label{fig:mad_det}
\end{figure}

A high value of BPCER10 renders the MAD system inconvenient for practical purposes. The BPCER10 was increased by pre-selection (i.e., MagFace and ArcFace) and by the morphing algorithm used (i.e., UBO morpher and NTNU morpher). The trend is illustrated in more detail in Fig~\ref{fig:mad_det}. Higher BPCER and MACER values were produced by the respective FRSs if pre-selection was performed according to our proposed architecture, and especially if it was performed using ArcFace or MagFace embeddings. This was consistent across different morphing algorithms.

While we have already shown the superiority of pre-selection over random pairing, we also observe large differnces in MAD depending on which FRS is used to extract embeddings for training and testing the D-MAD classifiers. BPCER10 values were about half in magnitude when MagFace was used for D-MAD, regardless of which FRS was used to extract embeddings for image pre-selection (Fig~\ref{fig:mad}). On the other hand, the advantage of attacks morphed by the UBO morpher over embeddings morphed by the NTNU morpher disappeared when MagFace was used for D-MAD compared to ArcFace (Fig~\ref{fig:mad}). The same can be seen in more detail in the DET curves (Fig~\ref{fig:mad_det}). The MACER and BPCER values were generally smaller, indicating a better performance of the D-MAD algorithm.

Interestingly, in some cases in Fig~\ref{fig:mad_det}, it can be observed that there was not a consistent bias of the D-MAD algorithms towards being fooled by morphs pre-selected by embeddings of the same FRS as the one used for D-MAD. MagFace embeddings for pre-selection performed best in fooling the D-MAD classifier in most cases, even if D-MAD was trained with ArcFace embeddings.

\section*{Discussion}
\label{discussion}

\subsection*{Comparison of face recognition models for pre-selection}
\label{sec:discussion:FRS}

Regarding the FRS for extracting embeddings for image pre-selection, several models were evaluated. The results showed that the recently published MagFace algorithm performed best in most cases -- depending on which metric was used to quantify vulnerability -- closely followed by ArcFace. VGG-Face and in particular DeepFace showed relatively weak performance for morphing pre-selection. However, all pre-selection methods improved the success of the Morphing Attacks compared to random pairing (Figs~\ref{fig:res:MMPMR_MORPH_AF_MF}, \ref{fig:MVR_UBO}, \ref{fig:ecdf_ubo} \& \ref{fig:ecdf_ubo_cots}, \ref{fig:res:MMPMR_MORPH_COTS}, Tables~\ref{table:riapar} \& \ref{table:RMMR_ranks}). 
In addition, a bias was observed such that if the same FRS was used for pre-selection and verification, the FRS was more susceptible to the resulting morphs (Figs~\ref{fig:res:MMPMR_MORPH_AF_MF} \& \ref{fig:ecdf_ubo}). However, when two COTS FRSs were used, this bias was mitigated and the hierarchical ranking between the pre-selection methods was still the same (Fig~\ref{fig:ecdf_ubo_cots}) \& \ref{fig:res:MMPMR_MORPH_COTS}).

A -- at first glance -- counterintuitive observation can be made when comparing Fig~\ref{fig:res:MMPMR_MORPH_AF_MF} and \ref{fig:res:MMPMR_MORPH_DF_VF}: While more accurate FRSs such as MagFace and ArcFace were quite vulnerable to Morphing Attacks, less accurate FRSs such as VGG-Face or DeepFace showed little vulnerability, since the prodAvgMMPMRs when verified with these FRSs mostly accumulated around $0$. This trend suggests that as FRSs generally improve, so that after calibration to an FMR of, say $0.1\%$, the FNMR becomes lower, these more accurate -- in terms of recognition -- FRSs become more vulnerable to Morphing Attacks. Earlier we called this phenomenon the \emph{morphing attack paradox}, and the effect is also nicely illustrated in \cite{ngan_frvt_2021}.

The key element is the decision threshold, located somewhere between the distributions of the mated comparison distances and the non-mated comparison distances (Fig~\ref{fig:ecdf_ubo}). As long as a considerable proportion of the values of the mated-morph comparison distances is located below the threshold towards the mated comparison distances, the FRS will be quite vulnerable. Adjusting the decision threshold toward the distribution of the mated comparison distances would reduce this vulnerability. Adjusting this decision threshold would be best possible in MagFace as a verification model, as the mated and morphed distributions show a small overlap (Fig~\ref{fig:ecdf_ubo}). However, for a model as good as ArcFace, as well as the two COTS FRSs, the distributions had significant overlap, impeding a simple solution via adjustment. Furthermore, by adjusting the decision threshold in the direction of the mated comparison distances, FMR would decrease, which generally makes the system more secure -- even against zero-effort impostor attacks. This in turn would inevitably increase the FNMR of the system, making it less convenient for particular practical purposes. At this point, it should be recalled, that the morphs used in this study were generated in an automated fashion. A real-world attacker would be able to invest time and resources into creating one single high-quality morph through manual intervention and various image post-processing steps. Comparison scores of such manually created morphs would be even more challenging to distinguish from mated comparisons, even when using MagFace for verification.

Furthermore, from the distribution of the prodAvgMMPMRs in Fig~\ref{fig:res:MMPMR_MORPH_AF_MF} and \ref{fig:res:MMPMR_MORPH_DF_VF} -- the high vulnerability of MagFace and ArcFace and the low vulnerability of VGG-Face and DeepFace -- some conclusions can be drawn on the results of the MAPs (Fig~\ref{fig:MVR_UBO}). In particular, the high values in the four leftmost columns of each MAP matrix are likely to derive from the more vulnerable MagFace and ArcFace FRSs, and the two highly vulnerable COTS FRSs. Analogously, the rather low values in the two rightmost columns are likely to be driven by the less vulnerable DeepFace and VGG-Face FRSs. This can also be observed in the MAPavg and G-MAP analyses. When removing these two verifying FRSs which showed high FNMRs (Table~\ref{table:thresholds}), the MAPavg and G-MAP values also increased substantially.

When correcting the mated morph rates for the FNMR of a verification FRS, as was done using the RMMR metric (Eq~\ref{eq:RIAPAR}, Table~\ref{table:riapar}), the general pattern persisted that a verification FRS was most vulnerable to morphs from image pairs pre-selected with the embeddings of the identical FRS. However, by ranking the RMMR row-wise and averaging across pre-selection methods and morphing algorithms (Table~\ref{table:RMMR_ranks}), the pattern manifested that MagFace was best suited for pre-selection among the FRSs tested. ArcFace followed MagFace, then VGG-Face, and lastly DeepFace. The poorest performance was constantly seen with randomly pre-selected morphs.

Further we want to emphasize that we morphed all images in the database. By only selecting a particular amount of pairs (e.g., $20\%$ with the smallest distances) for morphing, the vulnerability rates would be higher. Therefore, the reported vulnerabilities are expected to rather represent a lower bound.

In a preliminary analysis of a different face data set, we also investigated the potential of different distance (or similarity) metrics to apply in the proposed pre-selection architecture. However, morphed face images pre-selected based on Cosine distance yielded superior results \ref{fig:metrics}. This is not surprising as the algorithms are typically designed in such a way \cite{deng_arcface_2019, meng_magface_2021}. Therefore we used Cosine distance for pre-selection in the present study.

\subsection*{Evaluation of morphing algorithms}
\label{sec:discussion:morphingalgo}

A clear performance gap between the morphing algorithms is a common thread that runs through all of the analyses. Morphed images created by the UBO morpher, closely followed by those morphed by the NTNU morpher, performed best in fooling both FRSs (Fig~\ref{fig:res:MMPMR_MORPH_AF_MF} and Table~\ref{table:riapar}) and also D-MAD classifiers (Figs~\ref{fig:mad} \& \ref{fig:mad_det}). Morphs created by Alyssaq morpher and MIPGAN however performed worse in the current analyses.

What can be seen from Figs~\ref{fig:mad}~\&~\ref{fig:mad_det} is that the morphing algorithm deployed had a higher impact on the success of the D-MAD algorithm than the pre-selection. Similarly,  the success in terms of fooling the verification FRSs can be seen in Fig~\ref{fig:res:MMPMR_MORPH_AF_MF}, and by comparing the MAPs between morphers (Fig~\ref{fig:MVR_UBO}, \ref{fig:MVR_NTNU}, \ref{fig:MVR_ALYSSAQ}, \& \ref{fig:MVR_MIPGAN}). Alyssaq and MIPGAN morphers performed rather poorly at fooling the D-MAD algorithm, even with pre-selection applied. The reason for the Alyssaq morpher might for instance be the shape of the resulting morph (Fig~\ref{fig:examplemorphs}). The Alyssaq morpher returned morphs that were cropped at the face edges in a non-rectangular fashion (Fig~\ref{fig:examplemorphs}), and not projected back onto one of the original images' backgrounds. This has probably helped the D-MAD algorithm in its decision during both training and testing, although the classification was not performed on the raw images, but on the extracted face embeddings. Real-world attackers would not use such a morph, e.g., in a passport fraud scenario. Furthermore, MIPGAN produced rather blurry images (Fig~\ref{fig:examplemorphs}). In the original implementation of MIPGAN \cite{zhang_mipgan_2021, venkatesh_can_2020}, the morphs were of higher quality, but also the original images used for morphing were of higher image quality than those of the database used in the study at hand. Thus, the morphing in latent space in the present case may have dropped many facial characteristics that could have been helpful in facilitating a Morphing Attack.

\subsection*{MagFace improved Morphing Attack Detection}
\label{sec:discussion:MAD}

Instead of adjusting decision thresholds to counter Morphing Attacks, MAD algorithms could be inserted into a face verification process. The concept of the D-MAD algorithm used in the study at hand was introduced by \cite{scherhag_deep_2020} and learned to distinguish between the distribution of the differences between two bona fide images and the distribution of differences between morphs and bona fide images (Fig~\ref{fig:mad_pipeline}), all in the embedding space.

Testing on morph images derived from random pairing produced the lowest BPCER10 values, indicating the highest accuracy and therefore lowest vulnerability of the D-MAD algorithm towards these morphs (Fig~\ref{fig:mad}). Testing on the other morphs -- with pre-selection applied according to our proposed architecture -- increased the BPCER10 values. Thus, the greatest vulnerability of the D-MAD classifier was seen for morphs pre-selected by MagFace, then ArcFace, VGG-Face, and finally DeepFace. This was true regardless of whether the D-MAD classifier was trained and tested with ArcFace embeddings or with MagFace embeddings.

In fact, the D-MAD algorithm trained with MagFace embeddings showed considerably lower BPCER10 values, regardless of the type of pre-selection. Therefore, using MagFace instead of ArcFace could be a significant improvement to the D-MAD classifier proposed by Scherhag et al. \cite{scherhag_deep_2020}. Note that only the embeddings of the MagFace algorithm were used, not any additional quality metrics returned by the MagFace model. However, the quality of an image was still incorporated into the embeddings by the way the loss function was constructed. In MagFace's loss function, high-quality samples of an individual are pulled toward the center of the multidimensional distribution, while low-quality samples are pushed toward its boundaries  \cite{meng_magface_2021}. In other words, during the training of MagFace, the magnitude of the face embeddings was made proportional to the Cosine distance to the respective class (i.e., individuals) centers \cite{fu2021effect}. Therefore, having different image qualities for the bona fide images and the morphed images results in an easier separation of the two groups by the classifier since their positions in the $512$-dimensional embedding space are farther apart than the positions of two high-quality bona fide images.

The proposed D-MAD classifier based on MagFace embeddings was also submitted to the Face Recognition Vendor Test (FRVT) \cite{ngan_nistir_2022} and achieved good results in detecting high-quality morphed images (\ref{fig:frvt}). The FRVT MORPH report was created shortly before the initial submission of our manuscript. The D-MAD algorithms in \ref{fig:frvt} were evaluated on high-quality morphed images, created with commercial tools. The illustrated DET curve shows particularly low BPCER values of the MagFace D-MAD algorithm (\emph{hdamag}) in the area of low APCER (i.e., MACER) values. Moreover, it outperformed the similar algorithm which is based on ArcFace embeddings instead of MagFace embeddings (\emph{hdaarcface}) for MACER/APCER values from $0$ to $0.1$ (decimal fraction), an area of relevant security settings regarding Morphing Attacks. However, on low-quality images (i.e., Fig~4 in \cite{ngan_nistir_2022}), our classifier only outperformed the ArcFace classifier below an MACER/APCER of $0.02$ (decimal fraction).
Our classifier outperformed or underperformed compared to the ArcFace classifier depending on the face data set used for evaluation. In general, the ArcFace classifier performed better above MACER/APCER values of $0.1$ (decimal fraction). However, at lower MACER/APCER values, the MagFace classifier achieved lower BPCER values on several face data sets throughout the different processing tiers, i.e., morphed face image data sets of different quality, such as for example in the Visa Border or TWENTE data sets \cite{ngan_nistir_2022}. A detailed analysis of why the classifier performed better on some data sets and for specific MACER/APCER and BPCER is beyond the scope of the current study.

We further submitted the MagFace D-MAD algorithm to FVC-onGoing: on-line evaluation of fingerprint recognition algorithms \cite{dorizzi_fingerprint_2009}, in the section for Differential Morph Attack Detection \cite{raja_morphing_2020}. Among all algorithms tested on the DMAD-SOTAMD\_P\&S-1.0 benchmark, the D-MAD algorithm based on MagFace achieved the lowest BPCER10 values ($0.84$\%), and the second lowest BPCER20 values ($4.39$\%, \ref{fig:boep}). However, it achieved high BPCER100 values (i.e., $100$\%). The benchmark contained high-quality images of faces that were printed and scanned and captured with a frontal pose, natural expressions, and good lighting. See \cite{noauthor_fvc-ongoing_2023} for more details on how the algorithm performed on morphs from data subjects of different age groups or ethnicities and on morphs produced by different morphing algorithms, post-processing pipelines, and so forth.

One aim of large-scale image pre-selection based on embeddings was to evaluate a method for providing sufficiently large data sets of morphed face images for training MAD algorithms. Interestingly, a recent study showed that image pre-selection for training MAD algorithms could be done in the opposite way to the present study \cite{damer2019detect}. It was shown, that training morphing pairs with low similarity can improve the performance of the MAD algorithm \cite{damer2019detect}.

In the present study, separate D-MAD algorithms were trained on either ArcFace or MagFace embeddings. However, a fusion of the two may have constructive effects. In particular, it may be that the combination of both D-MAD algorithms would perform better than the D-MAD algorithm based on MagFace embeddings performed alone.

\section*{Conclusion}
\label{sec:disc:conclusion}

This study analyzed the use of face embeddings in image pre-selection and Morphing Attack Detection. MagFace and ArcFace embeddings were found to be effective for image pre-selection, as the resulting attacks posed a significant threat to modern FRS, especially COTS systems. MagFace outperformed ArcFace in many scenarios. Face embeddings from these models are highly suitable for pre-selection, such as for instance for automatically generating large databases of morphed faces. Similarly, morphed images pre-selected by MagFace, closely followed by ArcFace, posed a considerable threat to MAD algorithms by escaping detection. Lastly, MagFace differential embeddings were found to be particularly useful for attack detection, as they can improve the performance of a D-MAD algorithm. Taken together, the results underline the dual benefit of embeddings for both pre-selection and MAD, i.e. for the attacker and the defender.


%
%
%

\section*{Acknowledgments}
This work is supported by the European Union’s Horizon 2020 research and innovation program under grant agreement No 883356 (iMARS). Roman Kessler has received a scholarship by the National Research Center for Applied Cybersecurity (ATHENE). The authors would like to thank Haoyu Zhang for generating morphed images using the MIPGAN approach. They would also like to thank Daniel Fischer for converting the \emph{hdamag} prototype to the NIST FRVT MORPH api.

\bibliography{references}

\begin{thebibliography}{10}

\bibitem{kortli_face_2020}
Kortli Y, Jridi M, Al~Falou A, Atri M.
\newblock Face {Recognition} {Systems}: {A} {Survey}.
\newblock Sensors. 2020;20(2):342.
\newblock doi:{10.3390/s20020342}.

\bibitem{sanchez_del_rio_automated_2016}
Sanchez~del Rio J, Moctezuma D, Conde C, Martin~de Diego I, Cabello E.
\newblock Automated border control e-gates and facial recognition systems.
\newblock Computers \& Security. 2016;62:49--72.
\newblock doi:{10.1016/j.cose.2016.07.001}.

\bibitem{noauthor_regulation_2017}
Regulation ({EU}) 2017/2226 of the {European} {Parliament} and of the {Council} of 30 {November} 2017 establishing an {Entry}/{Exit} {System} ({EES}) to register entry and exit data and refusal of entry data of third-country nationals crossing the external borders of the {Member} {States} and determining the conditions for access to the {EES} for law enforcement purposes, and amending the {Convention} implementing the {Schengen} {Agreement} and {Regulations} ({EC}) {No} 767/2008 and ({EU}) {No} 1077/2011; 2017.

\bibitem{scherhag_face_2016}
Scherhag U, Rathgeb C, Merkle J, Breithaupt R, Busch C.
\newblock Face {Recognition} {Systems} {Under} {Morphing} {Attacks}: {A} {Survey}.
\newblock IEEE Access. 2016;7:23012--23026.
\newblock doi:{10.1109/ACCESS.2019.2899367}.

\bibitem{Ferrara2016}
Ferrara M, Franco A, Maltoni D.
\newblock On the effects of image alterations on face recognition accuracy.
\newblock In: Face recognition across the imaging spectrum; 2016. p. 195--222.
\newblock Available from: \url{https://doi.org/10.1007/978-3-319-28501-6₉}.

\bibitem{uncw_morph_nodate}
{UNCW}. {MORPH} {Non}-{Commercial} {Release} {Whitepaper}.;.
\newblock Available from: \url{http://people.uncw.edu/vetterr/MORPH-NonCommercial-Stats.pdf}.

\bibitem{torkar_morphing_2021}
Torkar M.
\newblock Morphing {Cases} in {Slovenia}; 2021.Available from: \url{https://eab.org/events/program/220}.

\bibitem{venkatesh_face_2021}
Venkatesh S, Ramachandra R, Raja K, Busch C.
\newblock Face {Morphing} {Attack} {Generation} \& {Detection}: {A} {Comprehensive} {Survey}.
\newblock IEEE Transactions on Technology and Society. 2021; p. 1--1.
\newblock doi:{10.1109/TTS.2021.3066254}.

\bibitem{ferrara_face_2021}
Ferrara M, Franco A, Maltoni D.
\newblock Face morphing detection in the presence of printing/scanning and heterogeneous image sources.
\newblock IET Biometrics. 2021;10(3):290--303.
\newblock doi:{10.1049/bme2.12021}.

\bibitem{tapia2021}
Tapia JE, Busch C.
\newblock Single morphing attack detection using feature selection and visualization based on mutual information.
\newblock IEEE access : practical innovations, open solutions. 2021;9:167628--167641.
\newblock doi:{10.1109/ACCESS.2021.3136485}.

\bibitem{scherhag_biometric_2017}
Scherhag U, Nautsch A, Rathgeb C, Gomez-Barrero M, Veldhuis RNJ, Spreeuwers L, et~al.
\newblock Biometric {Systems} under {Morphing} {Attacks}: {Assessment} of {Morphing} {Techniques} and {Vulnerability} {Reporting}.
\newblock In: 2017 {International} {Conference} of the {Biometrics} {Special} {Interest} {Group} ({BIOSIG}). Darmstadt, Germany: IEEE; 2017. p. 1--7.
\newblock Available from: \url{http://ieeexplore.ieee.org/document/8053499/}.

\bibitem{roettcher_finding_2020}
Roettcher A, Scherhag U, Busch C.
\newblock Finding the {Suitable} {Doppelgänger} for a {Face} {Morphing} {Attack}.
\newblock In: 2020 {IEEE} {International} {Joint} {Conference} on {Biometrics} ({IJCB}). Houston, TX, USA: IEEE; 2020. p. 1--7.
\newblock Available from: \url{https://ieeexplore.ieee.org/document/9304878/}.

\bibitem{raghavendra_face_2017}
Raghavendra R, Raja K, Venkatesh S, Busch C.
\newblock Face morphing versus face averaging: {Vulnerability} and detection.
\newblock In: 2017 {IEEE} {International} {Joint} {Conference} on {Biometrics} ({IJCB}). Denver, CO: IEEE; 2017. p. 555--563.
\newblock Available from: \url{http://ieeexplore.ieee.org/document/8272742/}.

\bibitem{raja_morphing_2020}
Raja K, Ferrara M, Franco A, Spreeuwers L, Batskos I, Gomez-Barrero FdWM, et~al.
\newblock Morphing {Attack} {Detection} -- {Database}, {Evaluation} {Platform} and {Benchmarking}.
\newblock arXiv:200606458 [cs]. 2020;.

\bibitem{amos_openface_nodate}
Amos B, Ludwiczuk B, Satyanarayanan M.
\newblock {OpenFace}: {A} general-purpose face recognition library with mobile applications.
\newblock CMU School of Computer Science; p.~20.

\bibitem{will_koehrsen_neural_2018}
{Will Koehrsen}. Neural {Network} {Embeddings} {Explained}.; 2018.
\newblock Available from: \url{https://towardsdatascience.com/neural-network-embeddings-explained-4d028e6f0526}.

\bibitem{shi_probabilistic_2019}
Shi Y, Jain A.
\newblock Probabilistic {Face} {Embeddings}.
\newblock In: 2019 {IEEE}/{CVF} {International} {Conference} on {Computer} {Vision} ({ICCV}). Seoul, Korea (South): IEEE; 2019. p. 6901--6910.
\newblock Available from: \url{https://ieeexplore.ieee.org/document/9008376/}.

\bibitem{schroff_facenet_2015}
Schroff F, Kalenichenko D, Philbin J.
\newblock {FaceNet}: {A} {Unified} {Embedding} for {Face} {Recognition} and {Clustering}.
\newblock 2015 IEEE Conference on Computer Vision and Pattern Recognition (CVPR). 2015; p. 815--823.
\newblock doi:{10.1109/CVPR.2015.7298682}.

\bibitem{deng_arcface_2019-1}
Deng J, Guo J, Xue N, Zafeiriou S.
\newblock {ArcFace}: {Additive} {Angular} {Margin} {Loss} for {Deep} {Face} {Recognition}.
\newblock arXiv:180107698 [cs]. 2019;.

\bibitem{meng_magface_2021}
Meng Q, Zhao S, Huang Z, Zhou F.
\newblock {MagFace}: {A} {Universal} {Representation} for {Face} {Recognition} and {Quality} {Assessment}.
\newblock Proceedings of the IEEE/CVF Conference on Computer Vision and Pattern Recognition. 2021; p.~10.

\bibitem{gomez-barrero_is_2017}
Gomez-Barrero M, Rathgeb C, Scherhag U, Busch C.
\newblock Is your biometric system robust to morphing attacks?
\newblock In: 2017 5th {International} {Workshop} on {Biometrics} and {Forensics} ({IWBF}). Coventry: IEEE; 2017. p. 1--6.
\newblock Available from: \url{https://ieeexplore.ieee.org/document/7935079/}.

\bibitem{ferraraMorphVulnerabilityRate2022}
Ferrara M, Franco A, Maltoni D, Busch C.
\newblock Morph {Attack} {Potential}.
\newblock proceedings of the IEEE International Workshop on Biometrics and Forensics (IWBF), Salzburg, Austria. 2022;.

\bibitem{isoiec_international_2023}
ISO/IEC. International {Standard} {ISO}/{IEC} 20059: {Methodologies} to evaluate the resistance of biometric recognition systems to morphing attacks; 2023.
\newblock Available from: \url{https://www.iso.org/standard/86084.html}.

\bibitem{venkatesh_influence_2020}
Venkatesh S, Raja K, Ramachandra R, Busch C.
\newblock On the {Influence} of {Ageing} on {Face} {Morph} {Attacks}: {Vulnerability} and {Detection}.
\newblock In: 2020 {IEEE} {International} {Joint} {Conference} on {Biometrics} ({IJCB}). Houston, TX, USA: IEEE; 2020. p. 1--10.
\newblock Available from: \url{https://ieeexplore.ieee.org/document/9304856/}.

\bibitem{scherhag_deep_2020}
Scherhag U, Rathgeb C, Merkle J, Busch C.
\newblock Deep {Face} {Representations} for {Differential} {Morphing} {Attack} {Detection}.
\newblock arXiv:200101202 [cs]. 2020;.

\bibitem{deng_arcface_2019}
Deng J, Guo J, Xue N, Zafeiriou S.
\newblock {ArcFace}: {Additive} {Angular} {Margin} {Loss} for {Deep} {Face} {Recognition}.
\newblock arXiv:180107698 [cs]. 2019;.

\bibitem{parkhi_deep_2015}
Parkhi OM, Vedaldi A, Zisserman A.
\newblock Deep {Face} {Recognition}.
\newblock In: Procedings of the {British} {Machine} {Vision} {Conference} 2015. Swansea: British Machine Vision Association; 2015. p. 41.1--41.12.
\newblock Available from: \url{http://www.bmva.org/bmvc/2015/papers/paper041/index.html}.

\bibitem{taigman_deepface_2014}
Taigman Y, Yang M, Ranzato M, Wolf L.
\newblock {DeepFace}: {Closing} the {Gap} to {Human}-{Level} {Performance} in {Face} {Verification}.
\newblock In: 2014 {IEEE} {Conference} on {Computer} {Vision} and {Pattern} {Recognition}. Columbus, OH, USA: IEEE; 2014. p. 1701--1708.
\newblock Available from: \url{https://ieeexplore.ieee.org/document/6909616}.

\bibitem{serengil2020lightface}
Serengil SI, Ozpinar A.
\newblock {LightFace}: {A} hybrid deep face recognition framework.
\newblock In: 2020 innovations in intelligent systems and applications conference ({ASYU}); 2020. p. 23--27.

\bibitem{nguyen2010cosine}
Nguyen HV, Bai L.
\newblock Cosine similarity metric learning for face verification.
\newblock In: Asian conference on computer vision; 2010. p. 709--720.

\bibitem{ricanek_morph_2006}
Ricanek K, Tesafaye T.
\newblock Morph: {A} longitudinal image database of normal adult age-progression.
\newblock In: 7th international conference on automatic face and gesture recognition ({FGR06}). IEEE; 2006. p. 341--345.

\bibitem{alyssa_quek_face_nodate}
Quek A. face morpher;.
\newblock Available from: \url{https://github.com/alyssaq/face_morpher}.

\bibitem{Ferrara2014}
Ferrara M, Franco A, Maltoni D.
\newblock The magic passport.
\newblock In: {IEEE} international joint conference on biometrics; 2014. p. 1--7.

\bibitem{Ferrara2018}
Ferrara M, Franco A, Maltoni D.
\newblock Face demorphing.
\newblock IEEE Transactions on Information Forensics and Security. 2018;13(4):1008--1017.
\newblock doi:{10.1109/TIFS.2017.2777340}.

\bibitem{Ferrara2019}
Ferrara M, Franco A, Maltoni D.
\newblock Decoupling texture blending and shape warping in face morphing.
\newblock In: 2019 international conference of the biometrics special interest group ({BIOSIG}); 2019. p. 1--5.

\bibitem{kazemi_one_2014}
Kazemi V, Sullivan J.
\newblock One millisecond face alignment with an ensemble of regression trees.
\newblock In: 2014 {IEEE} {Conference} on {Computer} {Vision} and {Pattern} {Recognition}. Columbus, OH: IEEE; 2014. p. 1867--1874.
\newblock Available from: \url{https://ieeexplore.ieee.org/document/6909637}.

\bibitem{zhang_mipgan_2021}
Zhang H, Venkatesh S, Ramachandra R, Raja K, Damer N, Busch C.
\newblock {MIPGAN} - {Generating} {Strong} and {High} {Quality} {Morphing} {Attacks} {Using} {Identity} {Prior} {Driven} {GAN}.
\newblock IEEE Transactions on Biometrics, Behavior, and Identity Science. 2021;3(3):365--383.
\newblock doi:{10.1109/TBIOM.2021.3072349}.

\bibitem{venkatesh_can_2020}
Venkatesh S, Zhang H, Ramachandra R, Raja K, Damer N, Busch C.
\newblock Can {GAN} generated morphs threaten face recognition systems equally as landmark based morphs?-vulnerability and detection.
\newblock In: 2020 8th {International} {Workshop} on {Biometrics} and {Forensics} ({IWBF}). IEEE; 2020. p. 1--6.

\bibitem{manuel_aguado_martinez_pyeer_nodate}
Martinez MA. {PyEER};.
\newblock Available from: \url{https://github.com/manuelaguadomtz/pyeer}.

\bibitem{frontex2015best}
{Frontex}. Best practice technical guidelines for automated border control ({ABC}) systems; 2015.

\bibitem{singh_deep_2023}
Singh JM, Ramachandra R. Deep {Composite} {Face} {Image} {Attacks}: {Generation}, {Vulnerability} and {Detection}; 2023.
\newblock Available from: \url{http://arxiv.org/abs/2211.11039}.

\bibitem{venkatesh_detecting_2020}
Venkatesh S, Ramachandra R, Raja K, Spreeuwers L, Veldhuis R, Busch C.
\newblock Detecting {Morphed} {Face} {Attacks} {Using} {Residual} {Noise} from {Deep} {Multi}-{Scale} {Context} {Aggregation} {Network}.
\newblock Proceedings of the IEEE/CVF Winter Conference on Applications of Computer Vision. 2020; p.~10.

\bibitem{isoiec_international_2017}
{ISO/IEC}. International {Standard} {ISO}/{IEC} 30107-3 ({First} edition) {Information} technology - {Biometric} presentation attack detection - {Part} 3: {Testing} and reporting.; 2017.
\newblock Available from: \url{https://www.iso.org/standard/67381.html}.

\bibitem{ngan_frvt_2021}
Ngan M, Grother P, Hanaoka K, Kuo J. {FRVT} {MORPH}: {Face} {Morphing} {Detection} {Evaluation}; 2021.
\newblock Available from: \url{https://eab.org/cgi-bin/dl.pl?/upload/documents/2026/Ngan-FRVT-MORPH-NBLAW-210303.pdf}.

\bibitem{fu2021effect}
Fu B, Spiller N, Chen C, Damer N.
\newblock The effect of face morphing on face image quality.
\newblock In: 2021 international conference of the biometrics special interest group ({BIOSIG}); 2021. p. 1--5.

\bibitem{ngan_nistir_2022}
Ngan M, Grother P, Hanaoka K, Kuo J. {NISTIR} 8292 {DRAFT} {SUPPLEMENT}: {Face} {Recognition} {Vendor} {Test} ({FRVT}) {Part} 4: {MORPH} - {Performance} of {Automated} {Face} {Morph} {Detection} ({July} 14, 2022); 2022.
\newblock Available from: \url{https://github.com/usnistgov/frvt/blob/nist-pages/reports/morph/frvt_morph_report_2022_07_14.pdf}.

\bibitem{dorizzi_fingerprint_2009}
Dorizzi B, Cappelli R, Ferrara M, Maio D, Maltoni D, Houmani N, et~al.
\newblock Fingerprint and on-line signature verification competitions at {ICB} 2009.
\newblock In: Advances in {Biometrics}: {Third} {International} {Conference}, {ICB} 2009, {Alghero}, {Italy}, {June} 2-5, 2009. {Proceedings} 3. Springer; 2009. p. 725--732.

\bibitem{noauthor_fvc-ongoing_2023}
{FVC}-{onGoing} {Report} on {Differential} {Morph} {Attack} {Detection}; 2023.
\newblock Available from: \url{https://biolab.csr.unibo.it/FvcOnGoing/UI/Form/AlgResult.aspx?algId=8570}.

\bibitem{damer2019detect}
Damer N, Saladie AM, Zienert S, Wainakh Y, Terhörst P, Kirchbuchner F, et~al.
\newblock To detect or not to detect: {The} right faces to morph.
\newblock In: 2019 international conference on biometrics ({ICB}); 2019. p. 1--8.

\bibitem{phillips_overview_2005}
Phillips PJ, Flynn PJ, Scruggs T, Bowyer KW, Chang J, Hoffman K, et~al.
\newblock Overview of the {Face} {Recognition} {Grand} {Challenge}.
\newblock In: 2005 {IEEE} {Computer} {Society} {Conference} on {Computer} {Vision} and {Pattern} {Recognition} ({CVPR}'05). vol.~1. San Diego, CA, USA: IEEE; 2005. p. 947--954.
\newblock Available from: \url{http://ieeexplore.ieee.org/document/1467368/}.

\bibitem{vinh_information_2009}
Vinh NX, Epps J, Bailey J.
\newblock Information theoretic measures for clusterings comparison: is a correction for chance necessary?
\newblock In: Proceedings of the 26th {Annual} {International} {Conference} on {Machine} {Learning} - {ICML} ‘09; 2009. p.~8.

\bibitem{bradley2000constrained}
Bradley PS, Bennett KP, Demiriz A.
\newblock Constrained k-means clustering.
\newblock Microsoft Research, Redmond. 2000;20(0):0.

\end{thebibliography}



\newpage
\setcounter{figure}{0} 
\renewcommand{\thefigure}{S\arabic{figure} Fig} 
\captionsetup{labelformat=plain} 
\section*{Supporting information}

\begin{figure}[!h]
 \centering
 \includegraphics[width=.9\linewidth, trim={0 0 0 0}]{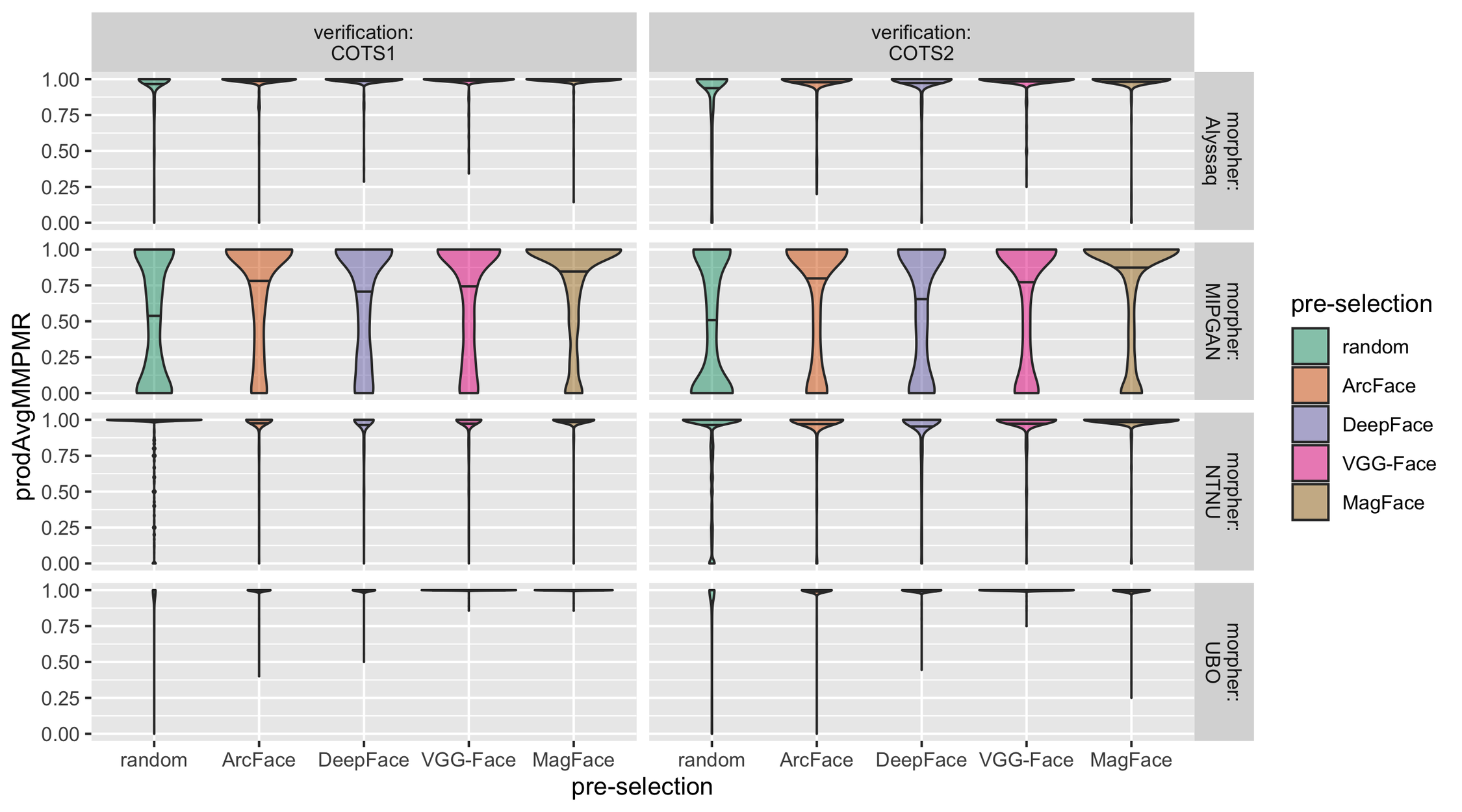} 
 \caption{{\bf Mated morphs comparison success rates for different image pre-selection embeddings.} 
All morphs have been evaluated by different COTS FRSs (columns). See Fig~\ref{fig:res:MMPMR_MORPH_AF_MF} for details.}
 \label{fig:res:MMPMR_MORPH_COTS}
\end{figure}

\begin{figure}[!h]
 \centering
 \includegraphics[width=.9\linewidth, trim={0 0 0 0}]{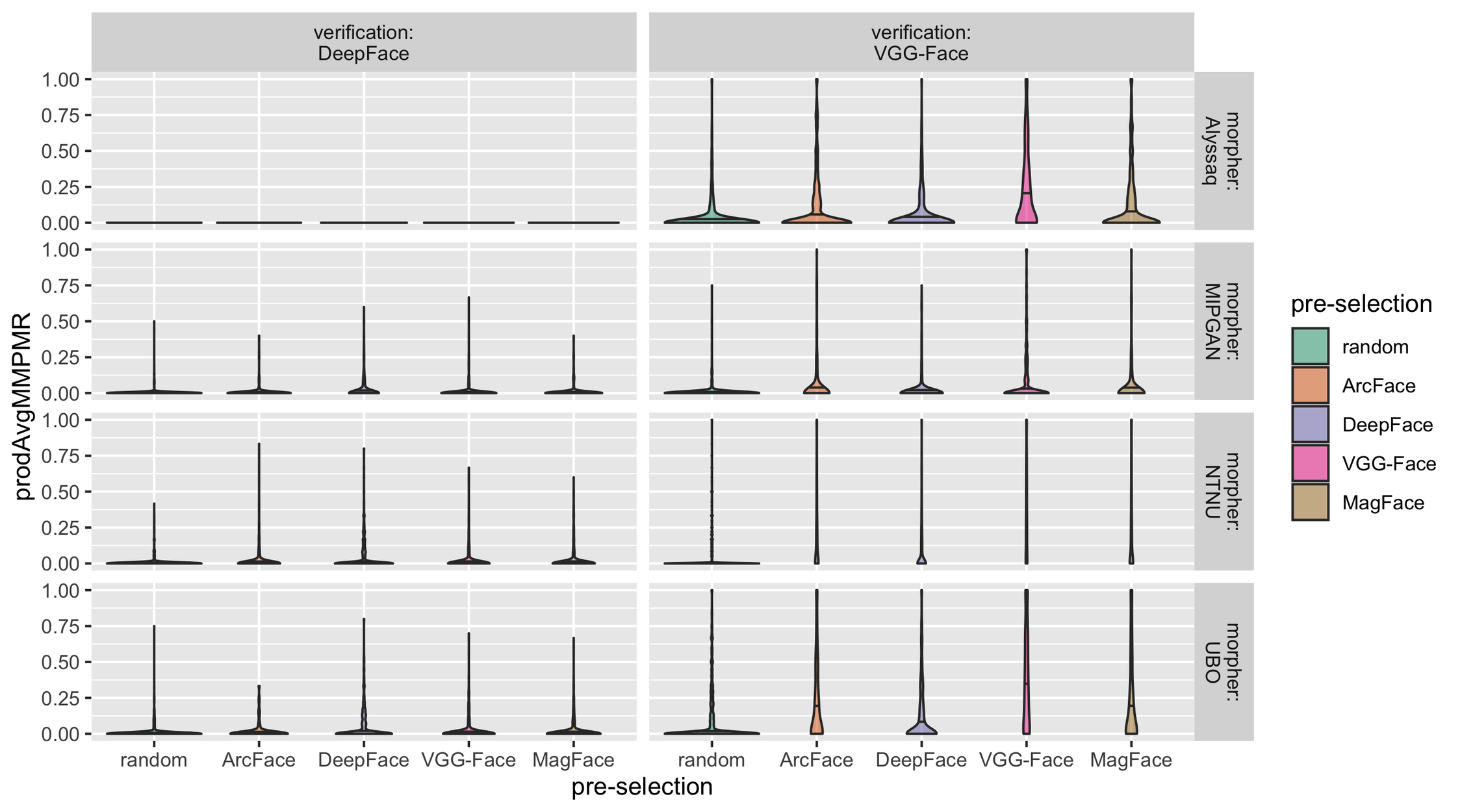} 
 \caption{{\bf Mated morphs comparison success rates for different image pre-selection embeddings.} 
All morphs have been verified using DeepFace and VGG-Face (columns). See Fig.~\ref{fig:res:MMPMR_MORPH_AF_MF} for verifications using ArcFace and MagFace and for more details.}
 \label{fig:res:MMPMR_MORPH_DF_VF}
\end{figure}

\begin{figure}[!h]
 \centering
 \includegraphics[width=.9\linewidth, trim={0 0 0 0}]{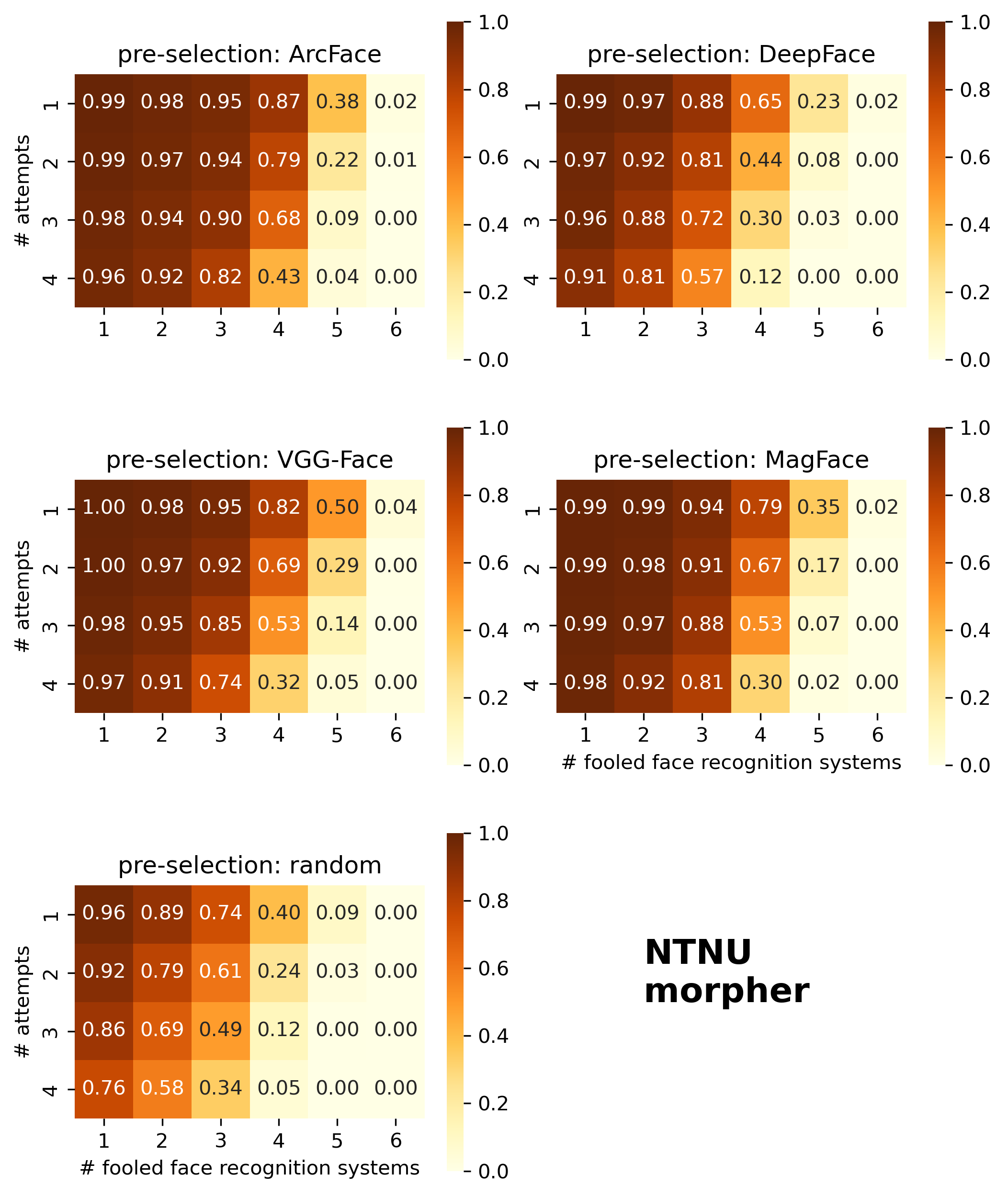} 
 \caption{{\bf MAP of morphs generated by NTNU morpher.}
 See Fig~\ref{fig:MVR_UBO} for details.}
 \label{fig:MVR_NTNU}
\end{figure}

\begin{figure}[!h]
 \centering
 \includegraphics[width=.9\linewidth, trim={0 0 0 0}]{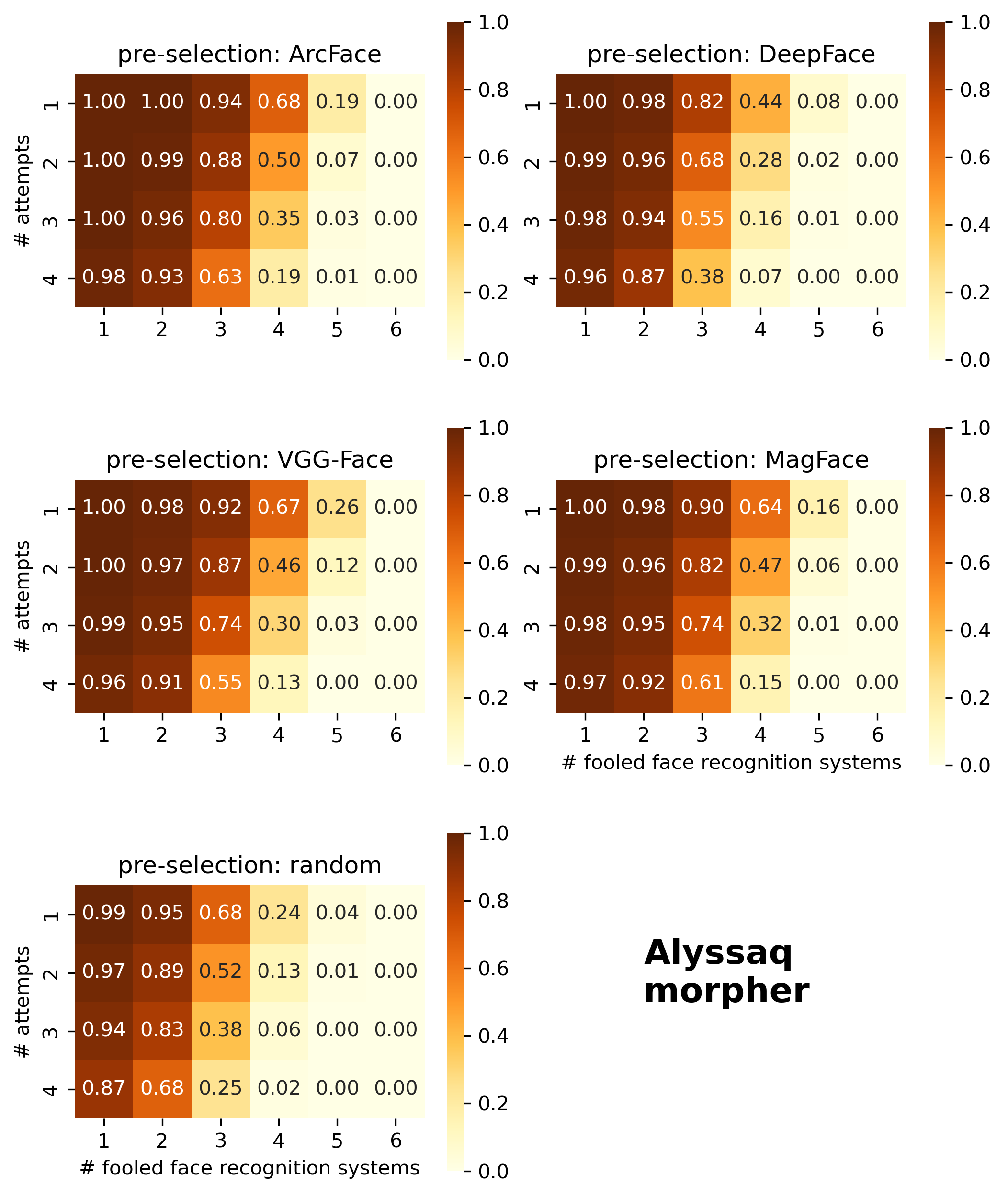} 
 \caption{{\bf MAP of morphs generated by Alyssaq morpher.}
 See Fig~\ref{fig:MVR_UBO} for details.}
 \label{fig:MVR_ALYSSAQ}
\end{figure}

\begin{figure}[!h]
 \centering
 \includegraphics[width=.9\linewidth, trim={0 0 0 0}]{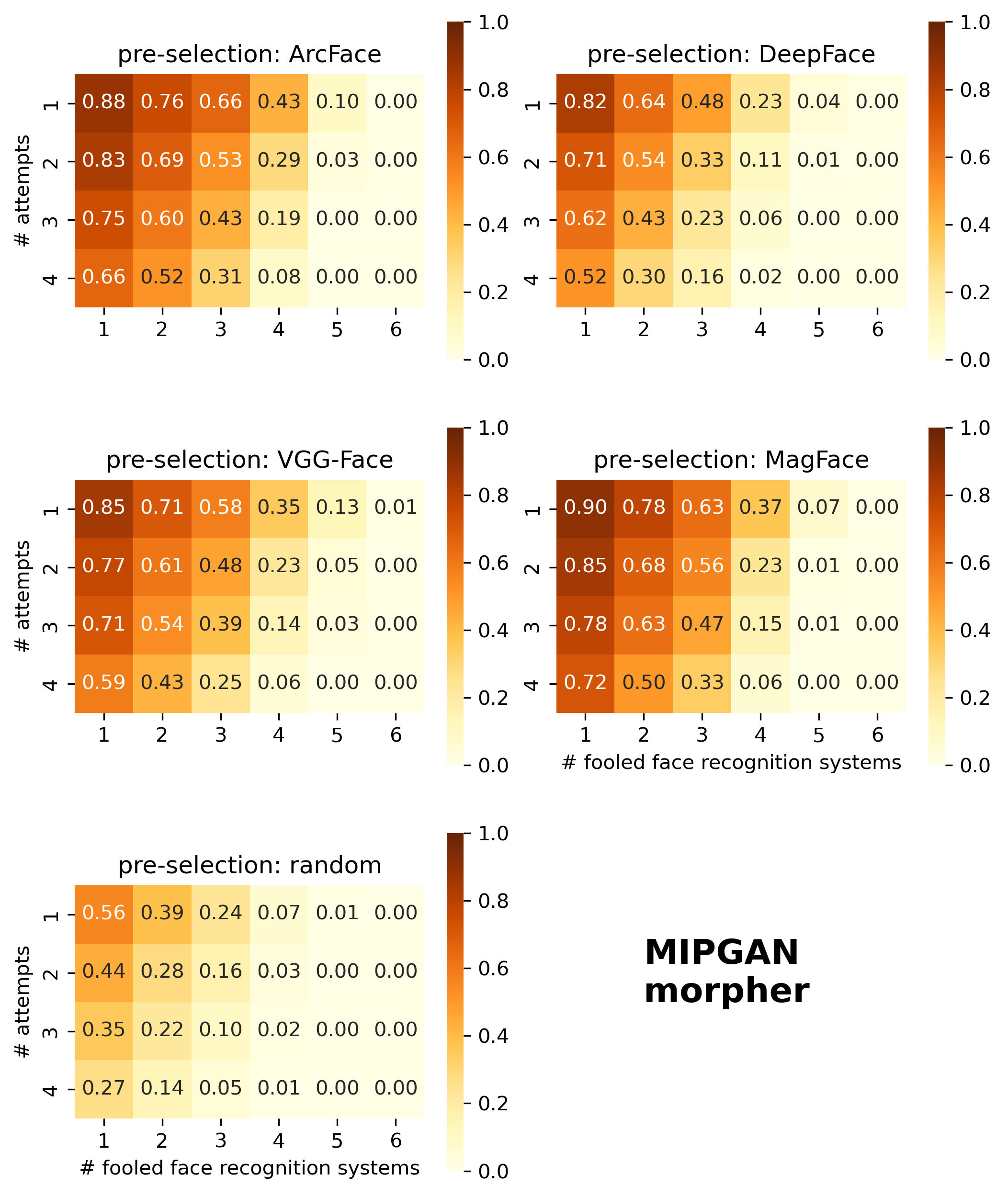} 
 \caption{{\bf MAP of morphs generated by MIPGAN morpher.}
 See Fig~\ref{fig:MVR_UBO} for details.}
 \label{fig:MVR_MIPGAN}
\end{figure}

\begin{figure}[!h]
 \centering
 \includegraphics[width=.9\linewidth, trim={0 0 0 0}]{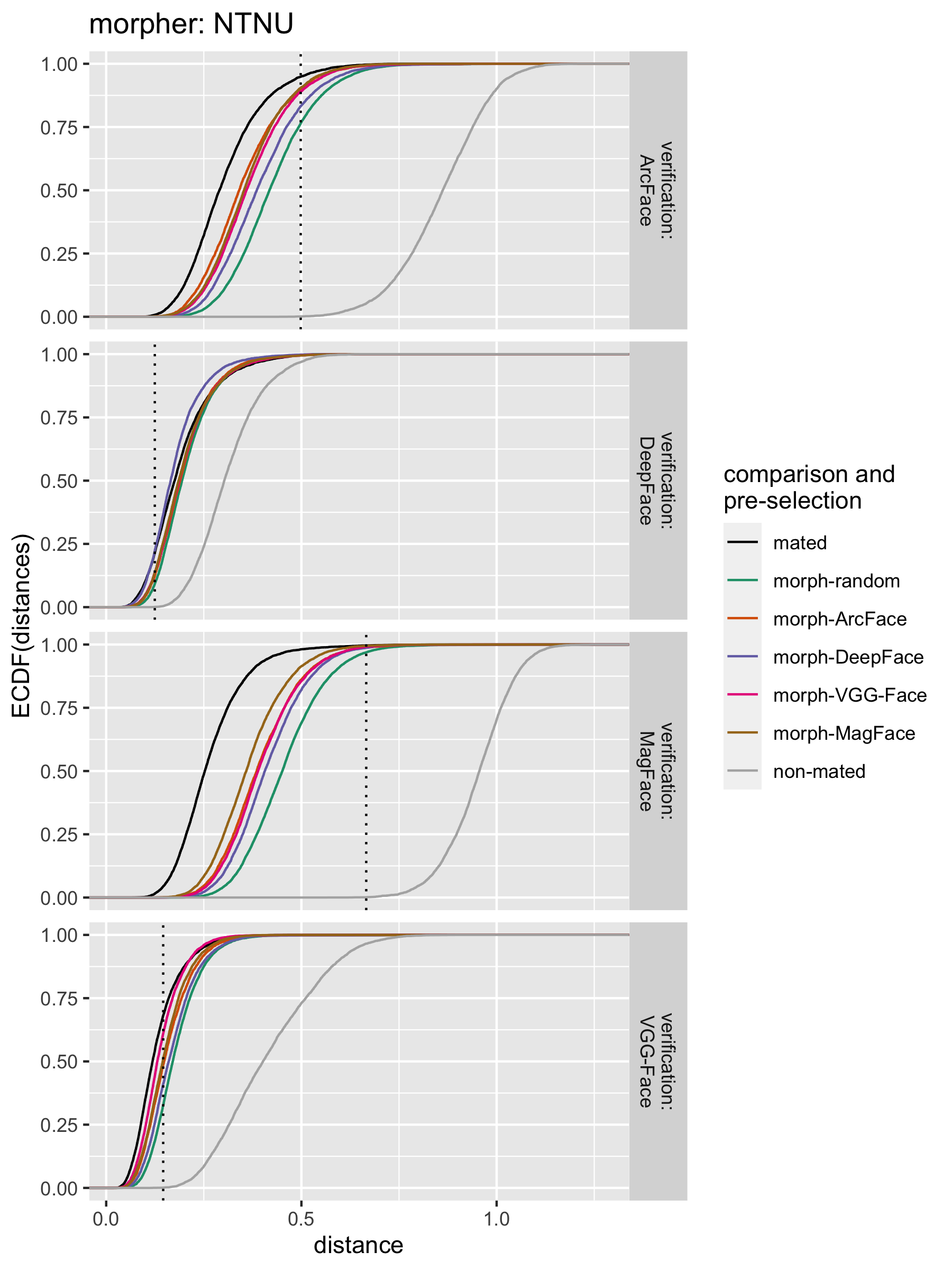} 
 \caption{{\bf ECDFs for distance scores of the open-source FRSs.}
 Morphs were created by NTNU morpher. See Fig~\ref{fig:ecdf_ubo} for details.}
 \label{fig:ecdf_ntnu}
\end{figure}

\begin{figure}[!h]
 \centering
 \includegraphics[width=.9\linewidth, trim={0 0 0 0}]{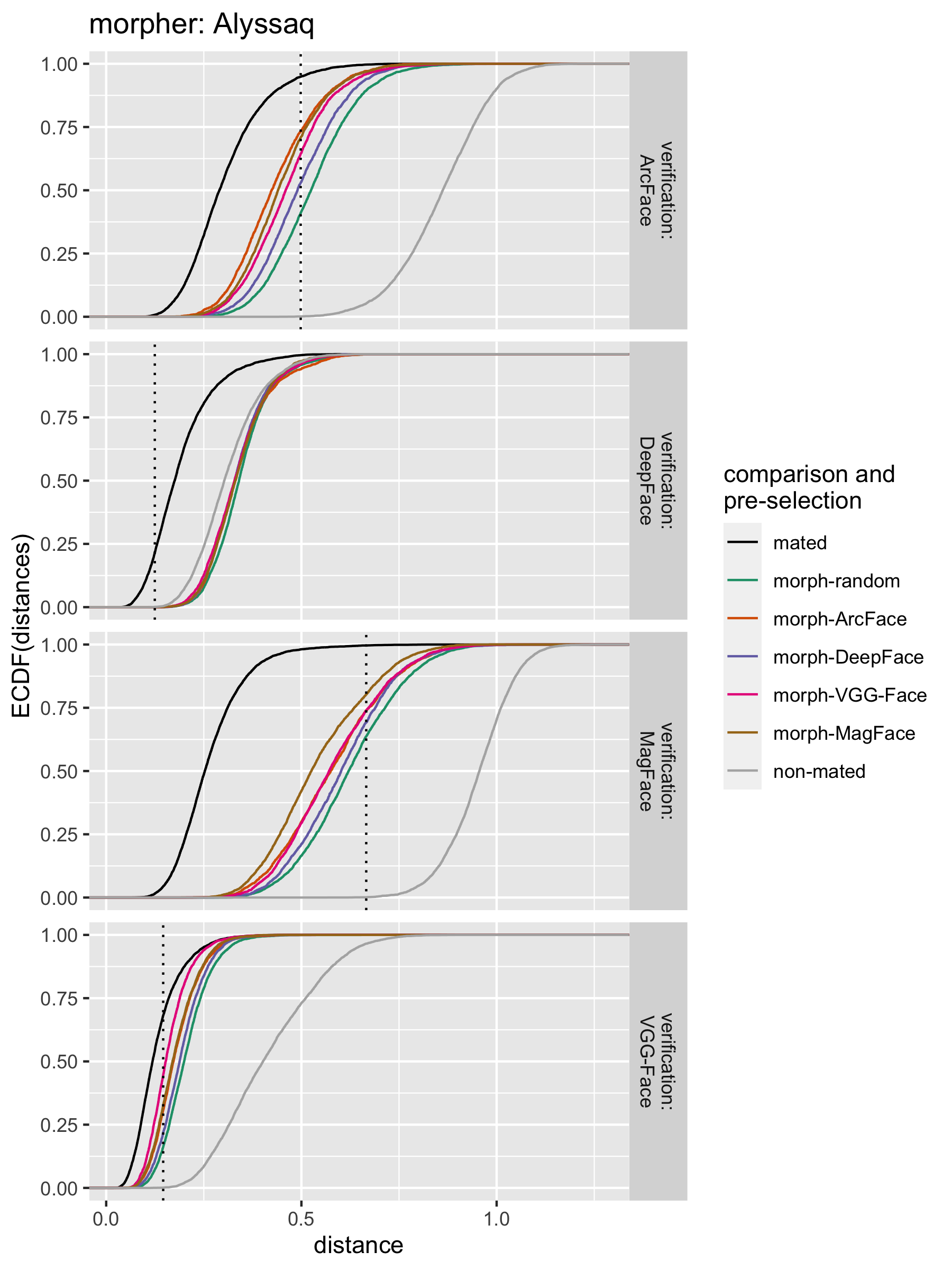} 
 \caption{{\bf ECDFs for distance scores of the open-source FRSs.}
 Morphs were created by Alyssaq morpher. See Fig~\ref{fig:ecdf_ubo} for details.}
 \label{fig:ecdf_alyssaq}
\end{figure}

\begin{figure}[!h]
 \centering
 \includegraphics[width=.9\linewidth, trim={0 0 0 0}]{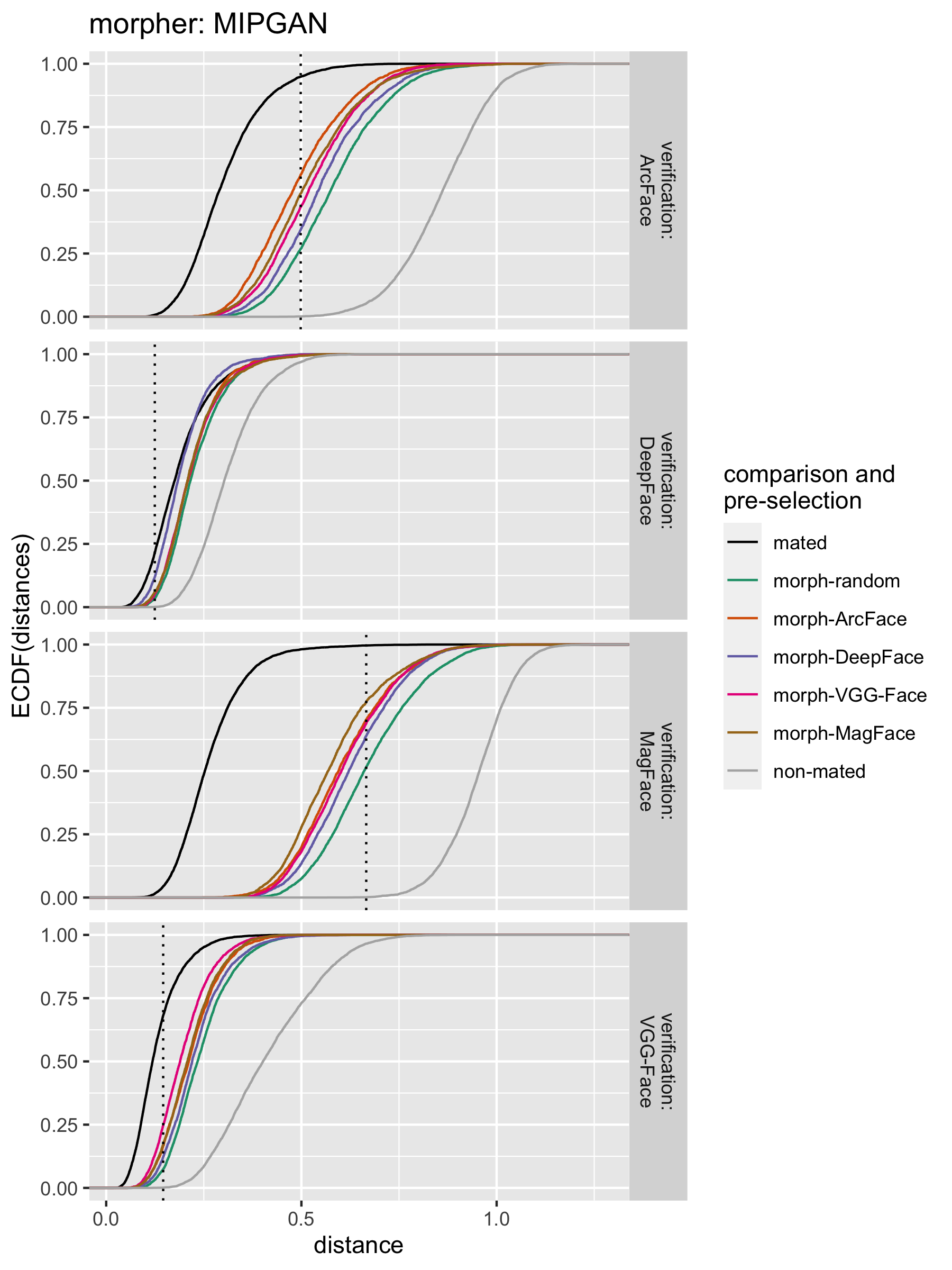} 
 \caption{{\bf ECDFs for distance scores of the open-source FRSs.}
 Morphs were created by MIPGAN morpher. See Fig~\ref{fig:ecdf_ubo} for details.}
 \label{fig:ecdf_mipgan}
\end{figure}

\begin{figure}[!h]
 \centering
 \includegraphics[width=.9\linewidth, trim={0 0 0 0}]{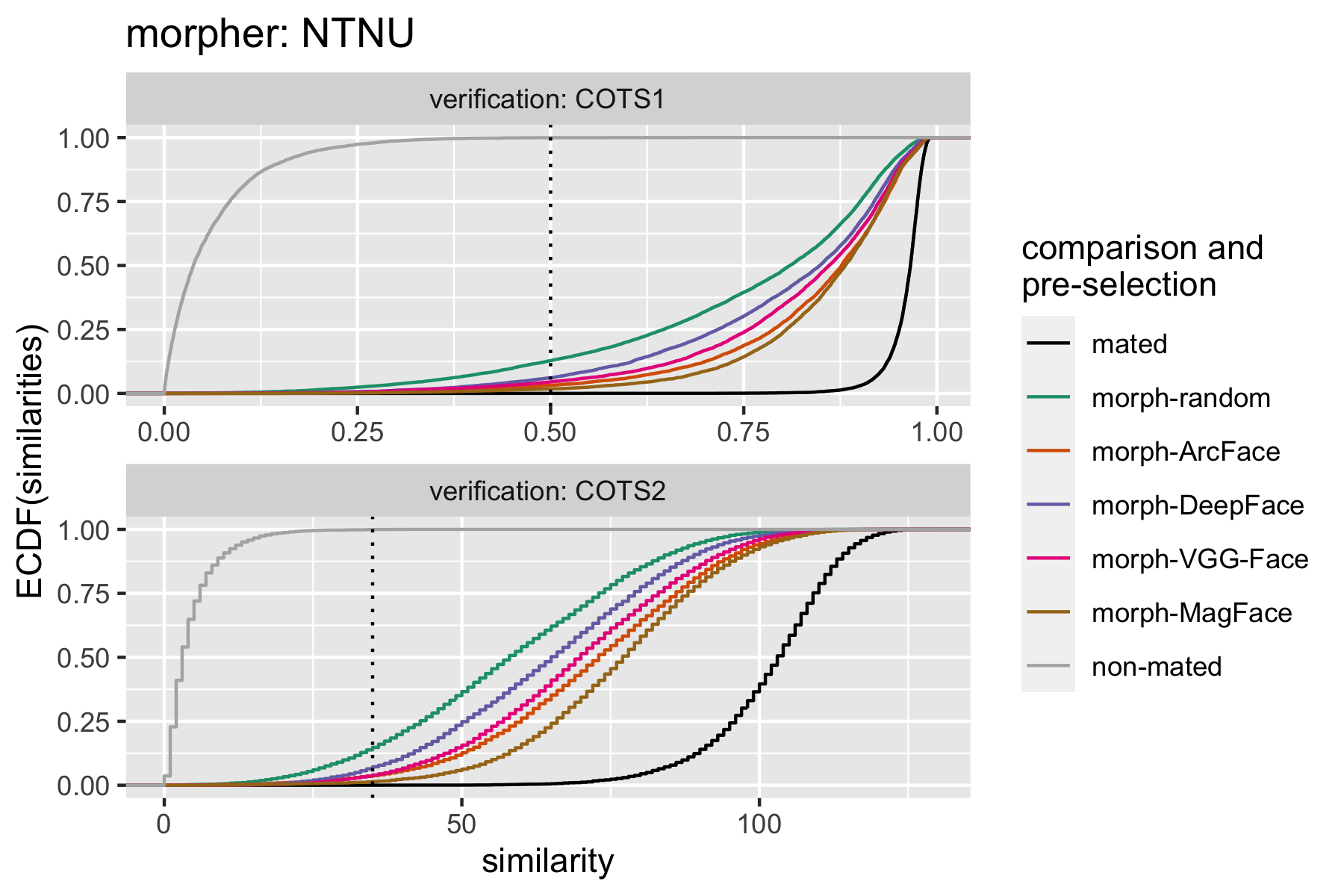} 
 \caption{{\bf ECDFs for similarity scores of the COTS FRSs.}
 Morphs were created by NTNU morpher. See Fig~\ref{fig:ecdf_ubo_cots} for details.}
 \label{fig:ecdf_ntnu_cots}
\end{figure}

\begin{figure}[!h]
 \centering
 \includegraphics[width=.9\linewidth, trim={0 0 0 0}]{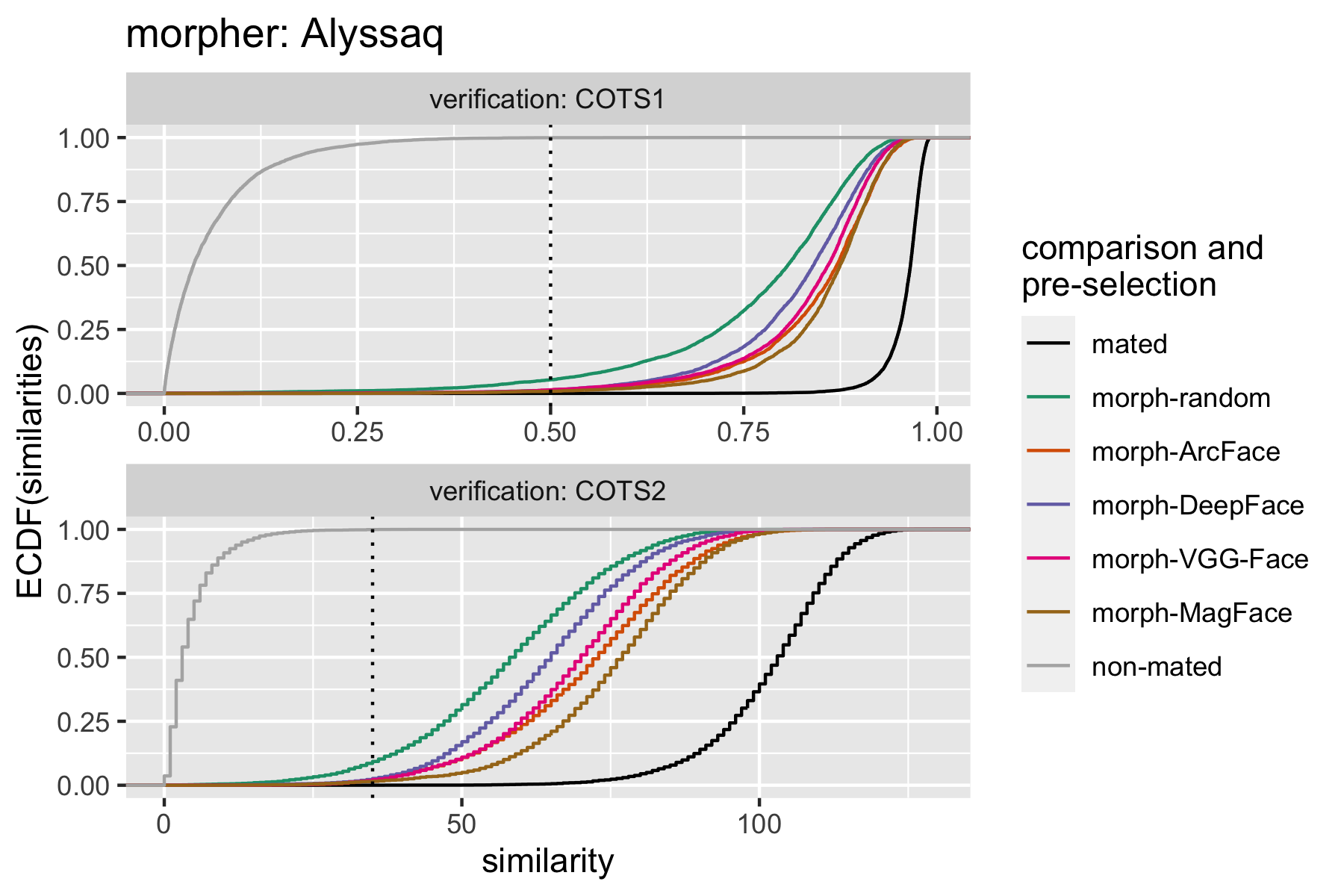} 
 \caption{{\bf ECDFs for similarity scores of the COTS FRSs.}
 Morphs were created by Alyssaq morpher. See Fig~\ref{fig:ecdf_ubo_cots} for details.}
 \label{fig:ecdf_alyssaq_cots}
\end{figure}

\begin{figure}[!h]
 \centering
 \includegraphics[width=.9\linewidth, trim={0 0 0 0}]{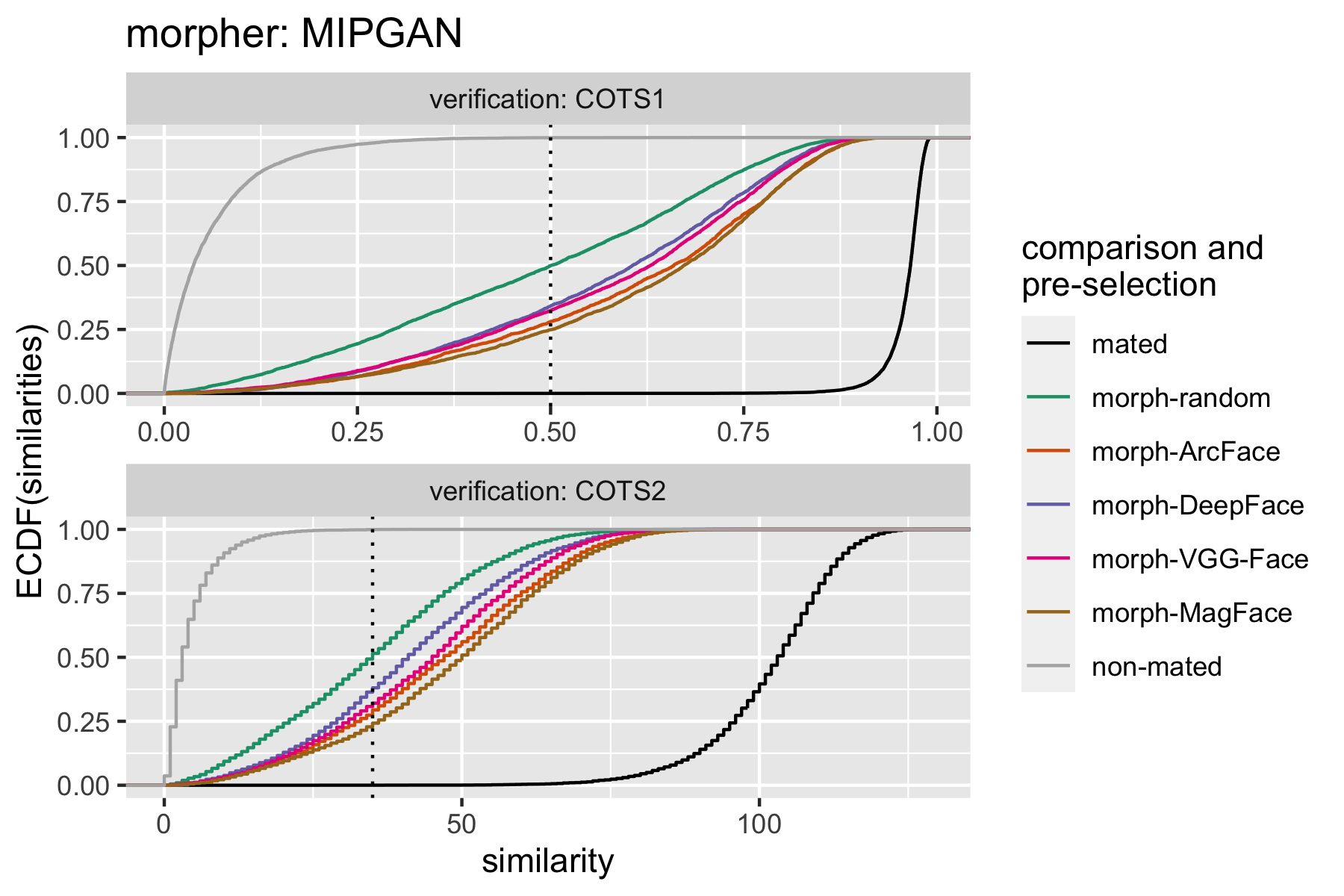} 
 \caption{{\bf ECDFs for similarity scores of the COTS FRSs.}
 Morphs were created by MIPGAN morpher. See Fig~\ref{fig:ecdf_ubo_cots} for details.}
 \label{fig:ecdf_mipgan_cots}
\end{figure}

\begin{figure}[!h]
 \centering
 \includegraphics[width=.9\linewidth, trim={0 0 0 0}]{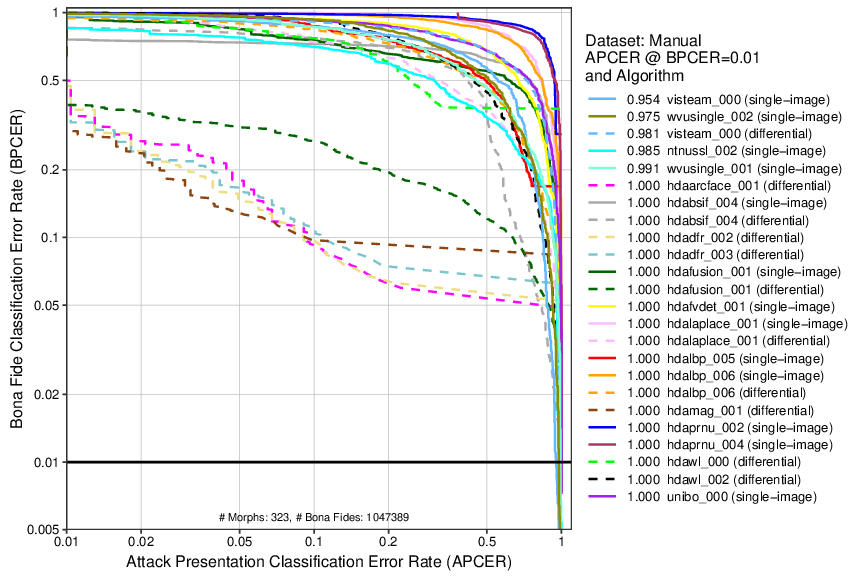} 
 \caption{{\bf FRVT-MORPH performance of the D-MAD algorithm using MagFace embeddings.} {\it Figure used with permission.}
Inspired by the superior performance of MagFace over ArcFace for D-MAD in the present study, the proposed algorithm using MagFace embeddings has also been submitted to the Face Recognition Vendor Test (FRVT) MORPH of the National Institute of Standards and Technology (NIST) \cite{ngan_nistir_2022}. In this test, the detection accuracy of the submitted classifier using MagFace embeddings for D-MAD (named \emph{hdamag}) was tested on morphed and bona fide face images of different data sets. Images of these data sets differed in various aspects, such as morph quality (automated vs. manually post-processed, publicly available tools, academic tools, or commercial tools) or with respect to the background, such as passport photographs vs. faces in natural environments. The figure illustrates the performance of the D-MAD algorithm on high-quality, manually post-processed morphed face images created with commercial morphing tools and high-quality, portrait-style bona fide images. The DET curves show good performance of the submitted \emph{hdamag} algorithm. In particular, it outperformed all other algorithms in the evaluation with the lowest BPCER at the relevant security settings of APCER (i.e., MACER) values below $0.1$.}
 \label{fig:frvt}
\end{figure}

\begin{figure}[!h]
 \centering
 \includegraphics[width=.9\linewidth, trim={0 0 0 0}]{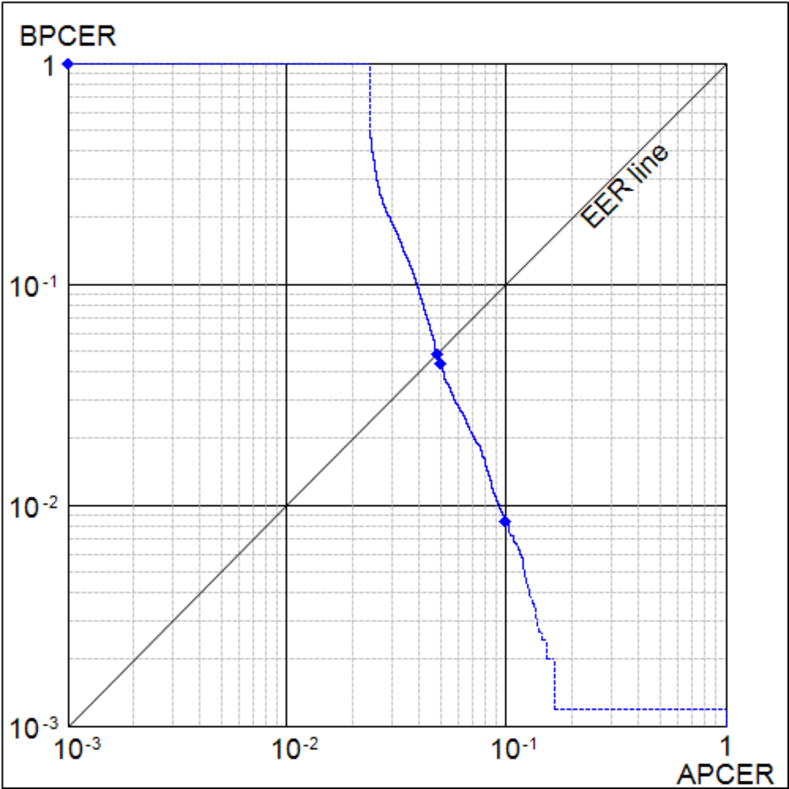} 
 \caption{{\bf FVC-ongoing performance of the D-MAD algorithm using MagFace embeddings.} {\it Figure used with permission.}
The MagFace D-MAD classifier has also been submitted to the FVC-onGoing section for Differential Morph Attack Detection \cite{dorizzi_fingerprint_2009, raja_morphing_2020}. The detection accuracy of the submitted algorithm using MagFace embeddings for D-MAD (named \emph{hdamag}) was tested on the DMAD-SOTAMD\_P\&S-1.0 benchmark. \emph{hdamag} achieved lowest BPCER10 and second lowest BPCER20 values \cite{noauthor_fvc-ongoing_2023}. The DET curve is illustrated in log scale, with BPCER10, BPCER20, EER, and BPCER100 illustrated as blue dots. }
 \label{fig:boep}
\end{figure}

\begin{figure}[!h]
 \centering
 \includegraphics[width=.9\linewidth, trim={0 0 0 0}]{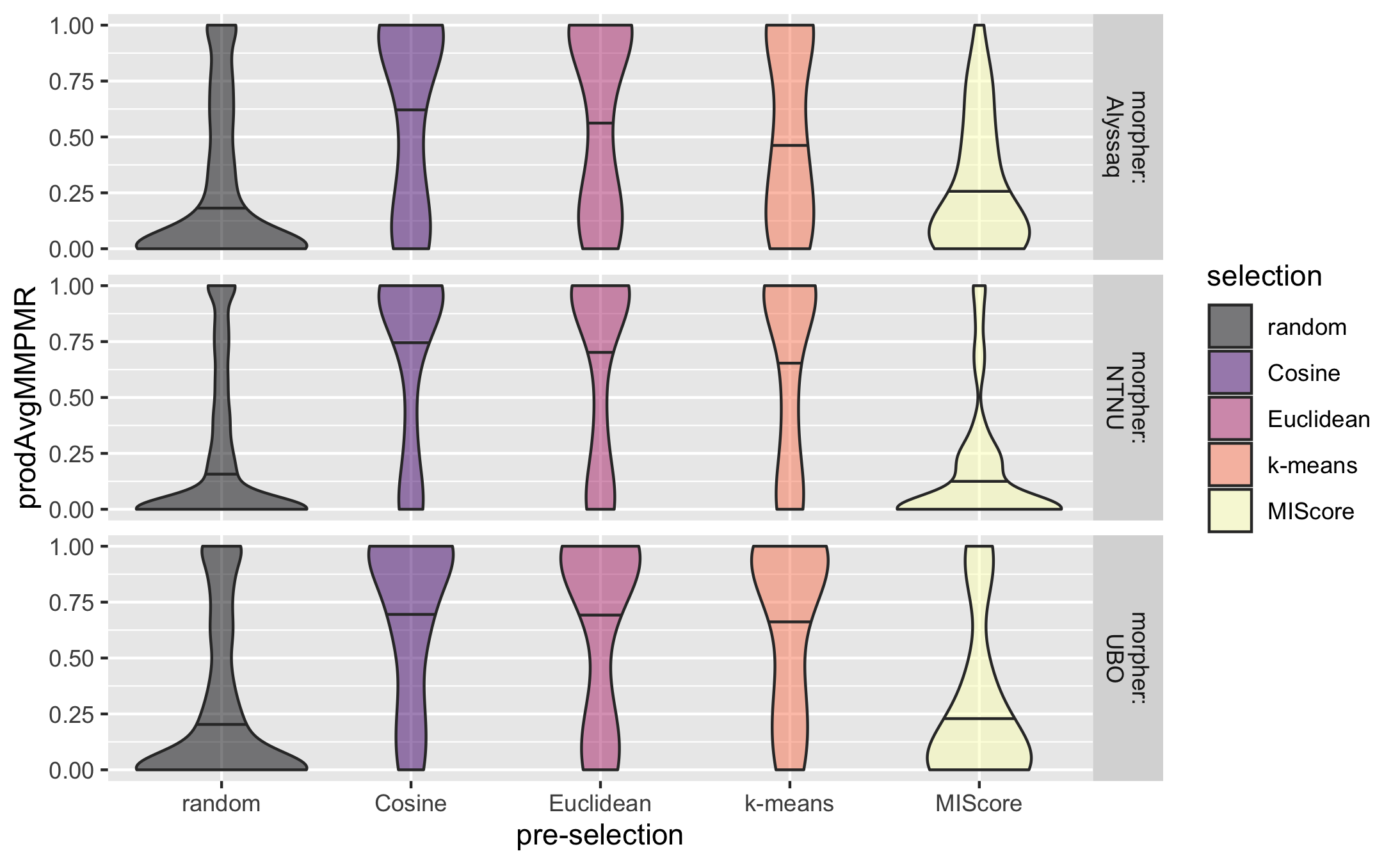} 
 \caption{{\bf Influence of the pre-selection distance or similarity metric on the attack potential.} 
In a preliminary study, we investigated different distance measures and their influence on the attack potential of the resulting morphs. Images from the Face Recognition Grand Challenge database version 2 (FRGCv2) data set \cite{phillips_overview_2005} were used for morphing. An in-house built subset of face images was used, comprising of 70 data subjects with around 10 captures each. We used different distance measures for pre-selection, such as Euclidean distance, Cosine distance, a distance measure based on the Mutual Information Score (miscore) \cite{vinh_information_2009}, or alternatively a clustering algorithm that inherently uses Euclidean distance (k-means) \cite{bradley2000constrained}. Morphs based on pre-selection using Cosine distance achieved the highest performance in terms of attack potential, closely followed by the two measures based on euclidean distance. For this reason, we continued with Cosine distance metric for the pre-selection algorithm in the present study.}
 \label{fig:metrics}
\end{figure}





\end{document}